\documentclass[10pt,technote,onecolumn]{IEEEtran}
\usepackage{cite}

\ifCLASSINFOpdf
   \usepackage[pdftex]{graphicx}
   \DeclareGraphicsExtensions{.pdf,.jpeg,.png}
\else
\fi
\usepackage{amsmath,amsthm,amssymb,amsfonts,bbm}
\newtheorem*{proof*}{Proof}
\newtheorem*{remark*}{Remark}

\usepackage{algorithmic,algorithm}

\usepackage{array}

\usepackage{multirow}
\usepackage{multicol}
\usepackage{color,xcolor}
\usepackage{setspace}
\usepackage{caption,subcaption}
\usepackage{booktabs}
\usepackage{diagbox}

\begin{document}
%
\title{Low-rank Tensor Grid for Image Completion}
%
%
%

\author{Huyan~Huang,
Yipeng~Liu,~\IEEEmembership{Senior Member,~IEEE},
Ce~Zhu,~\IEEEmembership{Fellow,~IEEE}
\thanks{This research is supported by National Natural Science Foundation of China (NSFC, No. 61602091, No. 61571102) and the Sichuan Science and Technology program  (No. 2019YFH0008, No.2018JY0035). The corresponding author is Yipeng Liu.}
\thanks{All the authors are with School of Information and Communication Engineering, University of Electronic Science and Technology of China (UESTC), Chengdu, 611731, China. (email: huyanhuang@gmail.com, yipengliu@uestc.edu.cn, eczhu@uestc.edu.cn).}
}
%
%

\markboth{Journal of \LaTeX\ Class Files,~Vol.~XX, No.~X, Month~Year}
{Shell \MakeLowercase{\textit{et al.}}: Bare Demo of IEEEtran.cls for IEEE Journals}
%



\maketitle

\begin{abstract}
Tensor completion estimates missing components by exploiting the low-rank structure of multi-way data. The recently proposed methods based on tensor train (TT) and tensor ring (TR) show better performance in image recovery than classical  ones. Compared with TT and TR, the projected entangled pair state (PEPS), which is also called tensor grid (TG), allows more interactions between different dimensions, and may lead to more compact representation. In this paper, we propose to perform image completion based on low-rank tensor grid. A two-stage density matrix renormalization group algorithm is used for initialization of TG decomposition, which consists of multiple TT decompositions. The latent TG factors can be alternatively obtained by solving alternating least squares problems. To further improve the computational efficiency, a multi-linear matrix factorization for low rank TG completion is developed by using parallel matrix factorization. Experimental results on synthetic data and real-world images show the proposed methods outperform the existing ones in terms of recovery accuracy.
\end{abstract}

\begin{IEEEkeywords}
projected entangled pair state, tensor grid, low rank tensor completion, alternating least squares, parallel matrix factorization.
\end{IEEEkeywords}

%
\IEEEpeerreviewmaketitle

\section{Introduction}
%
%
%
%

Tensors are generalizations of scalars, vectors, and matrices to multi-dimensional arrays with an arbitrary number of indices \cite{kolda2009tensor}. It is a natural representation for many multi-way images \cite{long2019low}. Tensor completion interpolates the missing entries by exploiting the multi-way data structures \cite{sidiropoulos2017tensor, cichocki2015tensor}, which has been used in a number of image recovery applications \cite{signoretto2011tensor, gandy2011tensor, duarte2011kronecker, liu2012tensor, liu2013tensor, kilmer2013third, mu2014square, du2016pltd, kasai2016online, bengua2017efficient, xiong2018field, he2019total}.

Most tensor completion methods are based on low rank assumptions induced from various decompositions. In tensor decomposition,  a higher-order tensor can be represented by a few of contracted lower-order tensors. 
CANDECOMP/PARAFAC (CP) decomposition factorizes a tensor into a linear combination of a series of rank-$1$ tensors \cite{kolda2009tensor}, and CP decomposition based completion methods have been studied in \cite{yang2015rank, liu2015trace, zhao2015bayesian, zhou2019tensor, liu2019low}.  Tucker (TK) decomposition  represents a tensor by a small core tensor multiplied by a number of matrices \cite{cichocki2015tensor, cichocki2016tensor}, and a number of TK decomposition based completion methods have been proposed in \cite{gandy2011tensor, xu2013parallel, liu2013tensor, liu2016generalized, yang2016iterative, liu2019low}. Besides, a recently widely used tensor singular value decomposition (t-SVD) factorizes a $3$-order tensor into two orthogonal tensors and a f-diagonal tensor based on t-product \cite{kilmer2013third}. The t-SVD based completion methods can refer to \cite{zhang2014novel, zhang2017exact}.

In addition to the aforementioned methods based on classical decompositions, there are a group of tensor networks based methods which show good performance in tensor completion, especially for higher-order data processing. For instance, using the tensor train (TT) decomposition which factorizes a tensor into a set of intermediate tensors and two border matrices \cite{oseledets2011tensor}, the methods proposed in \cite{wang2016tensor, bengua2017efficient} show advantage of high-order tensor completion compared with the above methods. However, the performance of TT decomposition based methods highly rely on the permutation of TT factors \cite{zhao2019learning}. The tensor ring (TR) decomposition represents a tensor as cyclically contracted $3$-order tensors \cite{zhao2019learning}, which is a linear combination of TT. The TR decomposition methods for tensor completion can refer to \cite{wang2017efficient, huang2019provable}.

Generally, the performance of a tensor completion method is highly affected by the used tensor decomposition. A suitable decomposition can well exploit the interactions between different tensor modes, though it is intractable to theoretically prove which tensor decomposition has the best representation ability. The recent TT decomposition and TR decomposition show better representation ability, which is verified by experiments \cite{bengua2017efficient, huang2019provable}.

As a two-dimensional generalization of TT, the projected entangled pair state (PEPS) can be regarded as a lattice by tensor network \cite{ye2018tensor}. One commonly used lattice of PEPS is square with open boundary conditions \cite{bridgeman2017hand}, thus we call the PEPS tensor grid (TG) decomposition. The TG has polynomial correlation decay with the separation distance \cite{bridgeman2017hand}, while TT and TR have exponential correlation decay. This indicates that PEPS has more powerful representation ability because it strengthens the interaction between different tensor modes. The tensor completion performance can be enhanced accordingly. We call a TG factor an activated tensor, and the corresponding environmental tensor is computed by contracting all other TG factors once a TG factor is assigned as an activated tensor.

Motivated by its outstanding representation capability, we use TG to exploit low-rank data structure for tensor completion in this paper. The proposed completion method interpolates the missing entries by minimizing the fitting error between the available data and their counterparts from computing a latent low-rank TG. Alternating least squares (ALS) method is used to solve the low-rank TG completion problem. The latent TG factors are optimized by solving a series of quadratic forms that consist of activated tensors and its environmental tensors. The single layer (SL) method \cite{pivzorn2011time} is used to accelerate the computation of environmental tensor. Besides, a more efficient parallel matrix factorization algorithm is used for low-rank TG completion too. In this algorithm, we consider the estimated tensor as a weighted sum of several tensors, each of which corresponding to a different TG unfolding. The recovery is performed by optimizing the parallel TG factors. The results of experiments on both synthetic data and real-world datasets show that TG structure significantly improve the representation ability compared with other structures.

The rest of this paper is organized as follows. Section \ref{section_notation} provides basic notations of tensor and preliminaries about tensor grid completion. The alternating least squares for low-rank tensor grid completion (ALS-TG) is presented in section \ref{section_algorithm}. In \ref{section_accelerated_algorithm}, we propose the parallel matrix factorization for low rank tensor grid completion (PMac-TG). The experimental results of synthetic data completion and real-world data completion are shown in section \ref{section_experiment}. The conclusion is drawn in section \ref{section_conclusion}.

\section{Notations and Preliminaries}
\label{section_notation}

\subsection{Notation of tensor}

Throughout this paper, a scalar, a vector, a matrix and a tensor are denoted by a normal letter, a boldfaced lower-case letter, a boldfaced upper-case letter and a calligraphic letter, respectively. For instance, a $D$-order tensor is denoted as $\mathcal{X} \in \mathbb{R}^{I_1 \times \dotsm \times I_D}$, where $I_d$ is the dimensional size corresponding to mode-$d$, $d \in \left\{1,\dotsc,D\right\}$.

The Frobenius norm of tensor $\mathcal{X}$ is defined as the squared root of the inner product of twofold tensors:
\begin{equation}
\lVert \mathcal{X} \rVert_\mathrm{F}=\sqrt{\langle \mathcal{X},\mathcal{X} \rangle}=\sqrt{\sum_{i_1=1}^{I_1}\cdots \sum_{i_D=1}^{I_D}{x_{i_1\cdots i_D}^2}}.
\end{equation}
The Hadamard product $\circledast$ is an element-wise product. For $D$-order tensors $\mathcal{X}$ and $\mathcal{Y}$, the representation is
\begin{equation}
\left(\mathcal{X} \circledast \mathcal{Y} \right)_{i_1\dotsm i_D}=x_{i_1\dotsm i_D}\cdot y_{i_1\dotsm i_D}.
\end{equation}

The mode-$d$ unfolding yields a matrix $\mathbf{X}^{\left(d\right)}\in \mathbb{R}^{I_d\times J_d}$ by unfolding $\mathcal{X}$ along its $d$-th mode, where $J_d=\prod^D_{n=1,\; n\neq d}I_n$.

\subsection{Preliminary on tensor grid decomposition}

\begin{figure}[htbp]
\centering
\begin{subfigure}[t]{0.2\textwidth}
\centering
\includegraphics[width=0.75\textwidth]{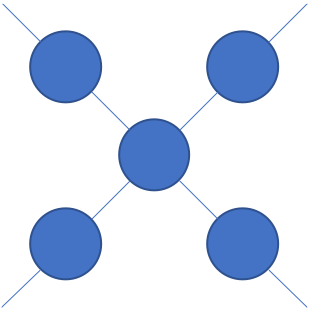}
\subcaption{A $5$-order CP/Tucker.}
\end{subfigure}
\begin{subfigure}[t]{0.2\textwidth}
\centering
\includegraphics[width=1\textwidth]{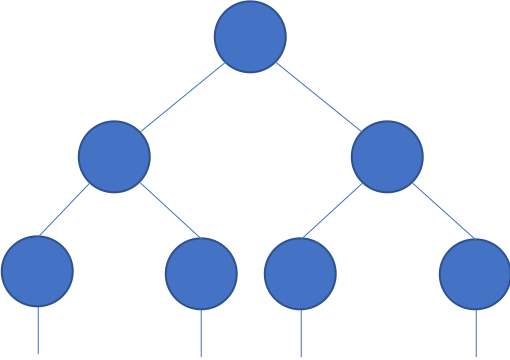}
\subcaption{A $4$-order HT.}
\end{subfigure}

\begin{subfigure}[t]{0.2\textwidth}
\centering
\includegraphics[width=1\textwidth]{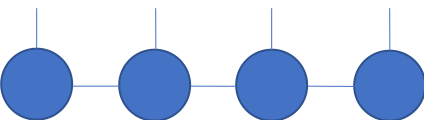}
\subcaption{A $4$-order TT.}
\end{subfigure}
\begin{subfigure}[t]{0.2\textwidth}
\centering
\includegraphics[width=1\textwidth]{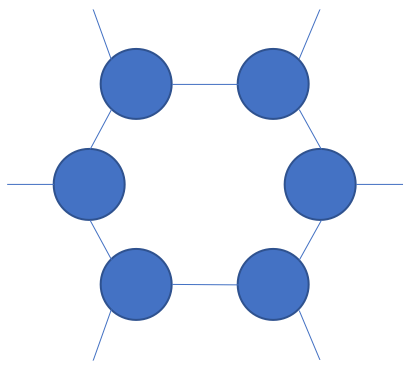}
\subcaption{A $6$-order TR.}
\end{subfigure}
\caption{An interpretation for the tensor network representations of CP/Tucker, HT, TT and TR decompositions. The circles represent factors and the lines mean the contractions of the connected factors.}
\label{tensor network}
\end{figure}

Tensor network is a powerful and convenient tool that contains enormous structures. By the aid of tensor network, the complicated mathematical expressions of many tensor decompositions can be represented by graphs. Fig. \ref{tensor network} exhibits the tensor network representations of CP/TK, hierarchical Tucker (HT), TT and TR decompositions. Specifically, CP and TK decompositions are star graphs, HT decomposition is a tree graph, TT decomposition is a path graph, and TR decomposition is a cycle graph.

In Fig. \ref{tensor network}, the circles represent factors, and the lines represent physical indices and virtual bonds. Each physical index corresponds to a tensor dimension. The virtual bonds denote contractions of the connected factors. The dimension of the virtual bonds is also called bond dimension $R$\cite{lubasch2014algorithms}. Conventionally, bond dimension is called rank in the context of tensor decomposition.

\begin{figure*}[ht]
\centering
\begin{subfigure}[t]{0.45\textwidth}
\centering
\includegraphics[scale=0.35]{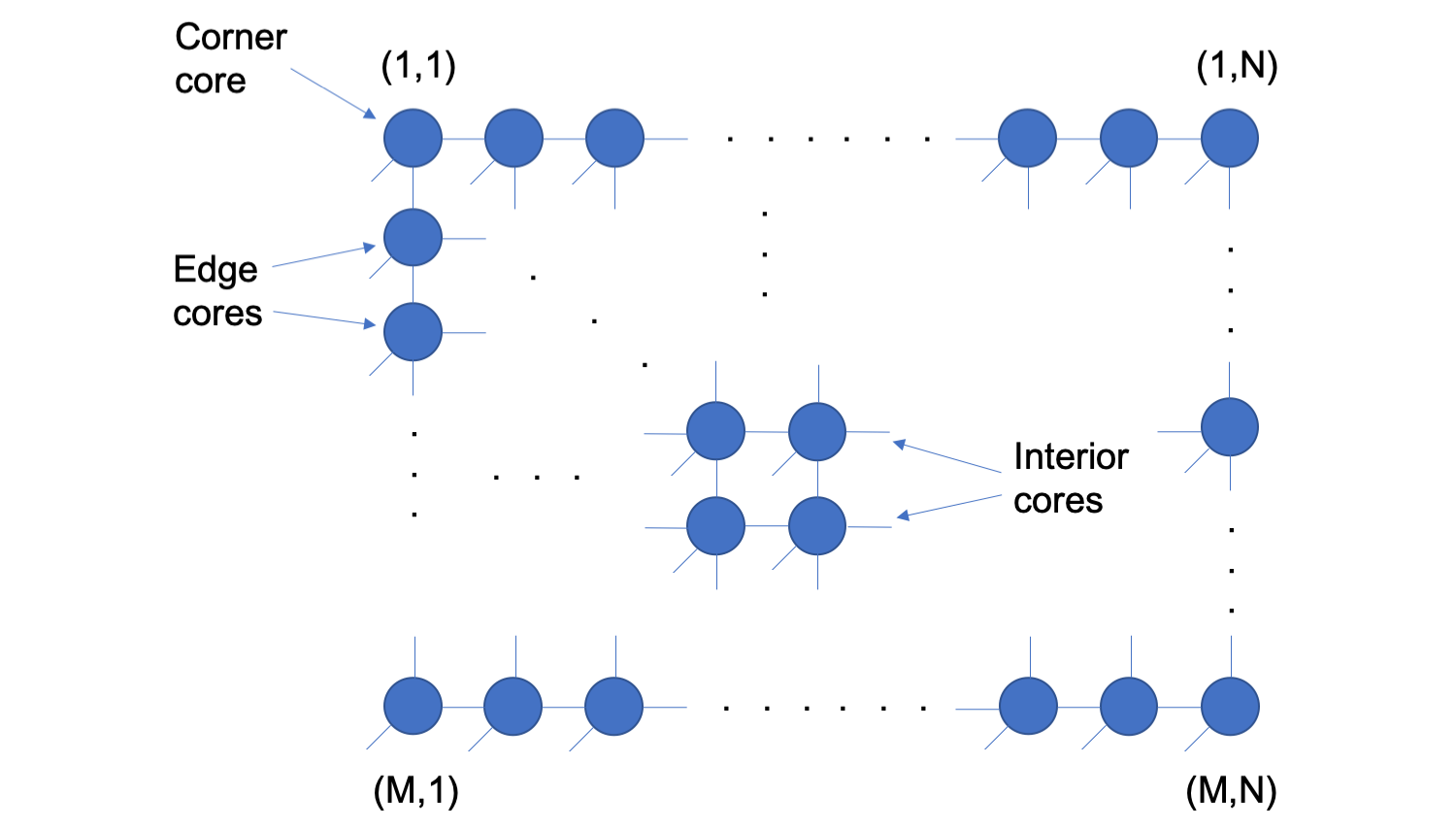}
\subcaption{The TG factors are arranged in a grid with locations varying from $\left(1,1\right)$ to $\left(M,N\right)$. According to the locations they are categorized into three types, i.e., corner, edge and interior factors.}
\end{subfigure}
\qquad \qquad
\begin{subfigure}[t]{0.45\textwidth}
\centering
\includegraphics[scale=0.35]{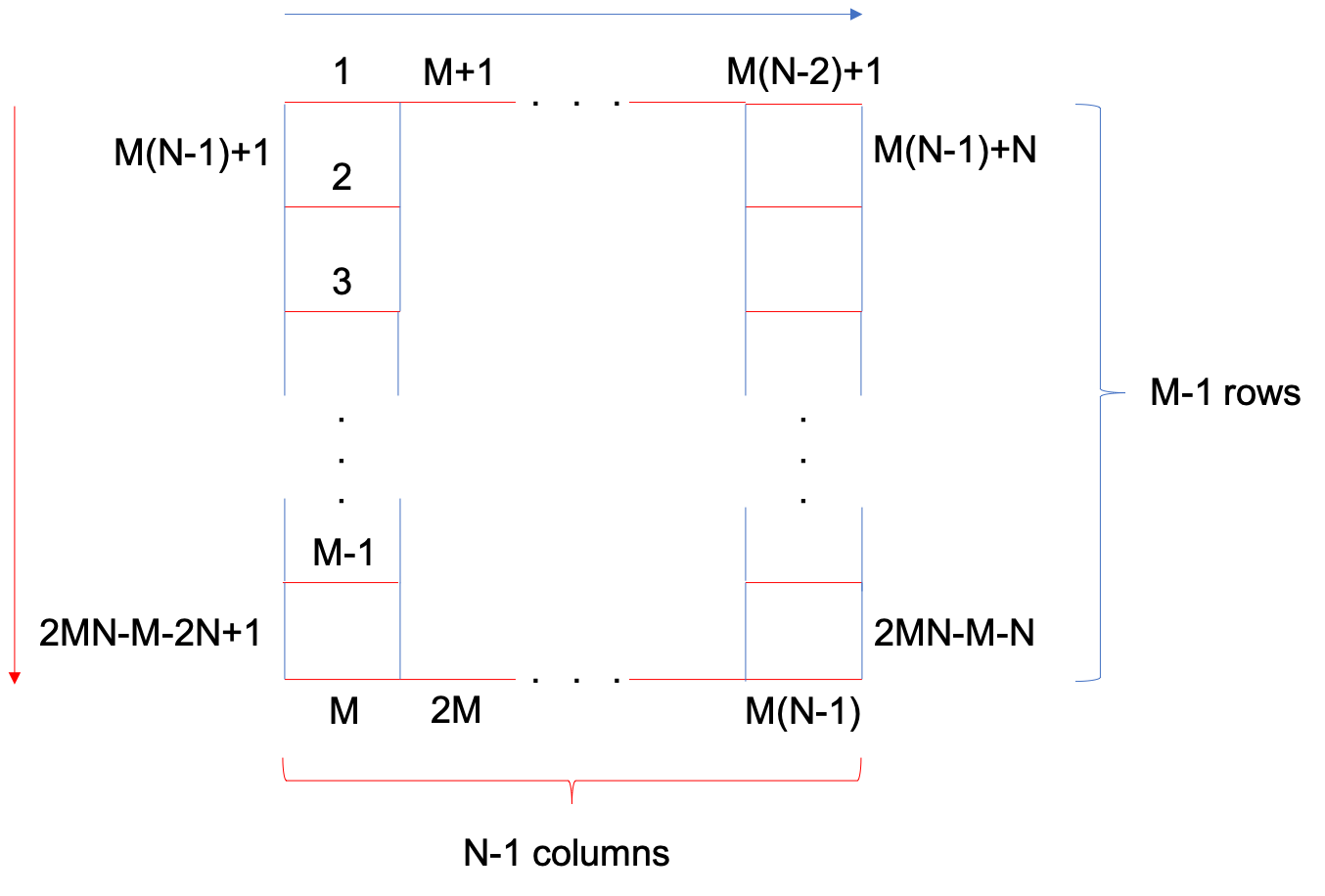}
\subcaption{The TG rank is divided into two classes, and we name they row and column ranks which are represented by the red lines and blue lines. The indices of the row ranks are $1,\; \dotsc,M\left(N-1\right)$, and the indices of the column ranks are $M\left(N-1\right)+1,\dotsc,2MN-M-N$.}
\end{subfigure}
\caption{A graphical illustration of TG decomposition.}
\label{TG decomposition1}
\end{figure*}

As it shows in Fig. \ref{TG decomposition1}(a), a TG is a Cartesian product of two path graphs using the tensor network representation \cite{ye2018tensor}. We divide the TG factors into three types, i.e., corner, edge and interior factors. For a TG of size $M\times N$, the TG factors are denoted by $\left\{\mathcal{C}^{\left(1,1\right)},\dotsc,\mathcal{C}^{\left(M,N\right)}\right\}$, where the superscript represents the positions of TG factors in the grid.

As shown in Fig. \ref{TG decomposition1}(b), we define the TG rank as a vector $\left[R_1,\dotsc,R_{2MN-M-N}\right]$, which is a concatenation of a row rank matrix $\mathbf{R}^{\text{R}}\in \mathbb{R}^{M\times N-1}$ and a column rank matrix $\mathbf{R}^{\text{C}}\in \mathbb{R}^{M-1\times N}$:
\begin{equation}
\label{thresholding of TG truncation}
\left[ R_1,\dotsc,R_{2MN-M-N} \right]=\left[ \operatorname{vec}\left(\mathbf{R}^{\text{C}}\right); \operatorname{vec}\left(\mathbf{R}^{\text{R}}\right) \right]^{\mathrm{T}}.
\end{equation}

We reshape the size vector $\left[I_1,\dotsc,I_{MN}\right]$ into a matrix $\mathbf{I}\in \mathbb{R}^{M\times N}$. There is no explicit definition of TG rank since TG contains cycles \cite{ye2018tensor, huang2019provable}. However, we can define the upper bonds for $\mathbf{R}^{\text{R}}$ and $\mathbf{R}^{\text{C}}$ as
\begin{equation}
\label{upper bond of TG-rank}
\left\{
\begin{aligned}
R^{\text{R}}_{mn}=& \min \left\{\prod^{n}_{j=1}I_{mj},\prod^{N}_{j=n+1}I_{mj}\right\} \\
R^{\text{C}}_{mn}=& \min \left\{\prod^{m}_{i=1}I_{in},\prod^{M}_{i=m+1}I_{in}\right\}
\end{aligned}
\right..
\end{equation}

Defining the indices
\begin{equation*}
\label{TG index}
\left\{
\begin{aligned}
& l_m=\left(n-2\right)M+m,\; r_m=\left(n-1\right)M+m \\
& u_{mn}=M\left(N-1\right)M+\left(m-2\right)N+n \\
& d_{mn}=M\left(N-1\right)M+\left(m-1\right)N+n \\
\end{aligned}
\right.,
\end{equation*}
then a TG factor can be denoted as $\mathcal{C}^{\left(m,n\right)}\in \mathbb{R}^{R_l\times R_r \times R_u \times R_d \times  I_{mn}}$.

For the tensor $\mathcal{X}$ of size $I_1\times \dotsm \times I_{MN}$ and TG rank $\left[R_1,\dotsc,R_{2MN-M-N}\right]$, the TG decomposition is represented as
\begin{align}
x_{i_1\dotsm i_{MN}}=& \sum_{m,n} \sum_{l_m,r_m,u_{mn},d_{mn}}c^{\left(1,1\right)}_{l_1,r_1,u_{11},d_{11},i_{11}}\dotsm \notag \\
& c^{\left(m,n\right)}_{l_m,r_m,u_{mn},d_{mn},i_{mn}} \dotsm c^{\left(M,N\right)}_{l_M,r_M,u_{MN},d_{MN},i_{MN}}.
\end{align}

\section{Alternating Least Squares for Tensor Grid Completion}
\label{section_algorithm}

The projection $\operatorname{P}_{\mathbb{O}}: \mathbb{R}^{I_1\times \dotsm \times I_D}\mapsto \mathbb{R}^{m}$ projects a tensor onto $\mathbb{O}$ to yield a vector, where
\begin{align*}
\mathbb{O}=\left\{ \left(i_1,\dotsc,i_D\right)|\text{ if entry }x_{i_1,\dotsc,i_D}\text{ is observed} \right\}.
\end{align*}
For example, the formulation for the $D$-order tensor $\mathcal{X}$ is 
\begin{equation}
\operatorname{P}_{\mathbb{O}}\left(\mathcal{X}\right)_{i_1\dotsm i_D}=
\left\{
\begin{aligned}
x_{i_1\dotsm i_D} \;\;\;& \left(i_1,\dotsc,i_D\right)\in \mathbb{O} \\
0 \;\;\;& \left(i_1,\dotsc,i_D\right)\notin \mathbb{O}
\end{aligned}
\right.
.
\end{equation}

Given a support set $\mathbb{O}$ for observations, we can assume the latent TG factors are $\left\{\mathcal{C}\right\}=\left\{\mathcal{C}^{\left(1\right)},\dotsc,\mathcal{C}^{\left(M,N\right)}\right\}$. The TG completion aims to recover the missing entries supported in $\mathbb{O}^{\perp}$ by optimizing the TG factors in a data fitting model as follows:
\begin{align}
\min_{\substack{\mathcal{C}^{\left(m,n\right)} \\ m=1,\dotsc,M \\ n=1,\dotsc,N}} \frac{1}{2}\lVert \operatorname{P}_{\mathbb{O}}\left(\operatorname{TG}\left(\left\{\mathcal{C}\right\}\right)\right)-\operatorname{P}_{\mathbb{O}}\left(\mathcal{T}\right)\rVert^2_2,
\label{TG completion model1}
\end{align}
where $\operatorname{TG}\left(\cdot\right)$ is an operator that  contracts the TG factors.

\subsection{Initialization}

All methods benefit from the initialized variables that are already close to the optimal solution \cite{lubasch2014algorithms}. Therefore, we try to develop an effective way for initializing the low-rank TG completion.

As we know, the density matrix renormalization group (DMRG) is used for efficient TT approximation \cite{oseledets2011tensor}. It reshapes the $D$-order tensor $\mathcal{X}$ of size $I_1\times \dotsm \times I_D$ into a matrix $\mathbf{X}^{\left(1\right)}\in \mathbb{R}^{I_1\times I_2\dotsm I_D}$, and applies the singular value decomposition (SVD) to the matrix, i.e., $\mathbf{X}^{\left(1\right)}=\mathbf{U}\boldsymbol{\Sigma}\mathbf{V}$. Then $\mathbf{U}\in \mathbb{R}^{I_1\times R_1}$ is reshaped as the first TT factor, and SVD is performed for the rest $\boldsymbol{\Sigma}\mathbf{V}$. The TT approximation is accomplished by performing $D-1$ SVD in such a way.

\begin{figure}
\centering
\includegraphics[scale=0.27]{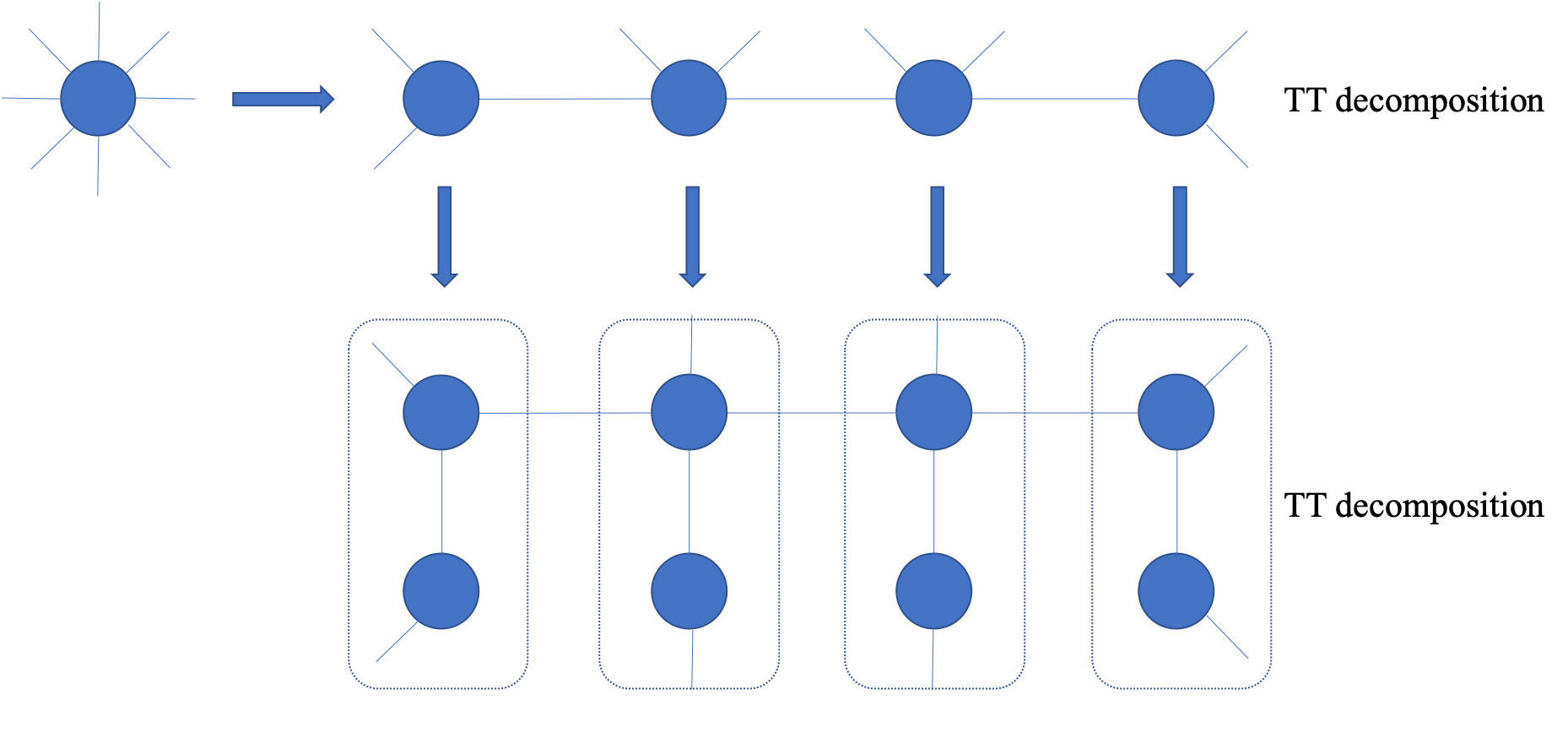}
\caption{The concept of TG decomposition. The SSVD method consists of two phases, each phase involves TT decomposition.}
\label{TG decomposition2}
\end{figure}

Our motivation comes from the TG contraction \cite{verstraete2008matrix, pivzorn2011time, lubasch2014algorithms}, in which it applies TT contraction to TG along one mode and results in a TT, then it  contracts this resultant TT to derive the tensor. We propose a spectral method called two-stage DMRG (2SDMRG) for efficient TG truncation, which is an inverse of computation of TG tensor. As it illustrates in Fig. \ref{TG decomposition2}, this method contains two phases. In the first phase, it applies DMRG to the tensor and derive a TT. In the second phase, it applies DMRG to each TT factor to derive the TG approximation. It is necessary to re-permute the orders of TT factors after the DMRG in the first phase. 

The key point in 2SDMRG is to choose the suitable truncated ranks, since each entry of the truncated TT rank derived by DMRG of the first phase is the product of $M$ row ranks (see the red lines in Fig. \ref{TG decomposition1}(b)) or $N$ column ranks (see the blue lines in Fig. \ref{TG decomposition1}(b)). A proper thresholding for TG truncation can be obtained by equation (\ref{upper bond of TG-rank}). The TG approximation is formed by extending the TG factors to the desired dimensions by filling i.i.d. Gaussian random variables.

The pseudocode of 2SDMRG for TG approximation is outlined in Algorithm \ref{algorithm_2SDMRG}. It can be applied to the zero-filled tensor for initialization of TG factors.
\begin{algorithm}
\caption{two-stage density matrix renormalization group (2SDMRG) for tensor grid approximation}
\label{algorithm_2SDMRG}
\begin{algorithmic}[1]
\REQUIRE Zero-filled tensor $\mathcal{T}$
\ENSURE The set of TG-factors $\left\{\mathcal{C}\right\}$
\STATE Compute the truncated TG-rank according to (\ref{upper bond of TG-rank})
\FOR{$j=1$ \TO $N$}
\STATE Apply DMRG to $\mathcal{T}$ to derive $\bar{\mathcal{C}}^{\left(j\right)}$
\ENDFOR
\FOR{$j=1$ \TO $N$}
\STATE Apply DMRG to $\bar{\mathcal{C}}^{\left(j\right)}$ to derive $\mathcal{C}^{\left(i,j\right)}$
\ENDFOR
\RETURN{$\left\{\mathcal{C}\right\}=\left\{ \mathcal{C}^{\left(1,1\right)},\dotsc,\mathcal{C}^{\left(M,N\right)} \right\}$}
\end{algorithmic}
\end{algorithm}

\subsection{Algorithm}

We first recall the bilinear form of matrix completion model. Supposing the ground truth matrix $\mathbf{M}\in\mathbb{R}^{I \times J}$ is can be factorized as $\mathbf{M}=\mathbf{A}\mathbf{B}$, where $\mathbf{A}\in\mathbb{R}^{I \times K}$ and $\mathbf{B}\in\mathbb{R}^{K \times J}$ are parallel factors of $\mathbf{C}$, and $K$ is the rank of $\mathbf{M}$. A non-convex optimization model for matrix completion can be formulated as follows:
\begin{equation*}
\label{matrix completion model}
\mathop{\min}_{\mathbf{A},\mathbf{B}} \frac{1}{2}\lVert \operatorname{P}_{\mathbb{O}}\left(\mathbf{A}\mathbf{B}\right)-\operatorname{P}_{\mathbb{O}}\left(\mathbf{M}\right)\rVert^2_{\mathrm{F}}.
\end{equation*}
The most widely used algorithm for this model is the alternating least squares, which alternately optimizes $\mathbf{A}$ and $\mathbf{B}$ with the other one being fixed.

Without loss of generality, we assume $\mathcal{C}^{\left(m,n\right)}$ is the activated tensor. We replace $\operatorname{P}_{\mathbb{O}}\left(\cdot\right)$ with $\mathcal{W}\circledast\left(\cdot\right)$, where $\mathcal{W}$ denotes a binary tensor in which a one-valued entry means the corresponding entry of $\mathcal{X}$ is observed and a zero-valued entry for the opposite. This model can be reformulated into a matrix form:
\begin{align}
\mathop{\min}_{\mathbf{A}^{\left(m,n\right)}} \frac{1}{2}\lVert \mathbf{W}^{\left(m,n\right)}\circledast\mathbf{A}^{\left(m,n\right)}\mathbf{B}^{\left(m,n\right)}-\mathbf{W}^{\left(m,n\right)}\circledast \mathbf{T}^{\left(m,n\right)} \rVert^2_{\mathrm{F}},
\label{sub-problem}
\end{align}
where $\mathbf{B}^{\left(m,n\right)}\in\mathbb{R}^{\hat{R}_{mn} \times  I_{\neq mn}}$ is the unfolding of the environmental tensor $\mathcal{B}^{\left(m,n\right)}$ derived by contracting all TG factors except the $\left(m,n\right)$-th one, $\mathbf{A}^{\left(m,n\right)}\in\mathbb{R}^{I_{mn}\times \hat{R}_{mn}}$ is the unfolding of the activated tensor $\mathcal{C}^{\left(m,n\right)}$, $\mathbf{W}^{\left(m,n\right)}\in\mathbb{R}^{I_{mn}\times I_{\neq mn}}$ and $\mathbf{T}^{\left(m,n\right)}\in\mathbb{R}^{I_{mn}\times I_{\neq mn}}$ are the unfoldings of $\mathcal{W}$ and $\mathcal{T}$, respectively, in which $I_{\neq mn}=\left(\prod^{M}_{i=1}\prod^{N}_{j=1}I_{ij}\right)/I_{mn}$.

The sub-problem (\ref{sub-problem}) can be divided into $I_{mn}$ sub-sub-problems with respect to the variables $\mathbf{a}^{\left(m,n\right)}_{i_{mn}}$, $i_{mn}=1,\dotsc,I_{mn}$, which is the $i_{mn}$-th row of $\mathbf{A}^{\left(m,n\right)}$. The corresponding model is
\begin{align}
\min_{\mathbf{a}^{\left(m,n\right)}_{i_{mn}}} \frac{1}{2}\lVert \mathbf{w}^{\left(m,n\right)}_{i_{mn}}\circledast \mathbf{a}^{\left(m,n\right)}_{i_{mn}}\mathbf{B}^{\left(m,n\right)}-\mathbf{w}^{\left(m,n\right)}_{i_{mn}}\circledast \mathbf{t}^{\left(m,n\right)}_{i_{mn}} \rVert^2_2,
\label{sub-sub-problem}
\end{align}
where $\mathbf{w}^{\left(m,n\right)}_{i_{mn}}$ is the $i_{mn}$-th row of $\mathbf{W}^{\left(m,n\right)}$ and $\mathbf{t}^{\left(m,n\right)}_{i_{mn}}$ is the $i_{mn}$-th row of $\mathbf{T}^{\left(m,n\right)}$.

To solve (\ref{sub-sub-problem}), we define a permutation matrix $\mathbf{P}^{\left(m,n\right)}_{i_{mn}}=\mathbf{e}_{\mathbb{D}^{\left(m,n\right)}_{i_{mn}}}\in \mathbb{R}^{I_{\neq mn}\times ||\mathbf{w}^{\left(m,n\right)}_{i_{mn}}||_0}$, where $\mathbf{e}_k$ is a vector of length $I_{\neq mn}$ whose values are all zero but one for the $k$-th entry, $k\in \mathbb{D}^{\left(m,n\right)}_{i_{mn}}$ and $\mathbb{D}^{\left(m,n\right)}_{i_{mn}}=\left\{j_{mn}|w^{\left(m,n\right)}_{i_{mn}j_{mn}}=1\right\}$. Expanding the squared term in (\ref{sub-sub-problem}), the solution to (\ref{sub-sub-problem}) is
\begin{align}
\mathbf{a}^{\left(m,n\right)^*}_{i_{mn}}=-\mathbf{g}^{\left(m,n\right)}_{i_{mn}}\mathbf{H}^{\left(m,n\right)^{\dagger}}_{i_{mn}},
\label{update}
\end{align}
where $\dagger$ is the Moore-Penrose pseudoinverse and
\begin{equation*}
\left\{
\begin{aligned}
& \mathbf{H}^{\left(m,n\right)}_{i_{mn}}=\widetilde{\mathbf{B}}^{\left(m,n\right)}_{i_{mn}}\widetilde{\mathbf{B}}^{{\left(m,n\right)}^{\mathrm{T}}}_{i_{mn}},\; \mathbf{g}^{\left(m,n\right)}_{i_{mn}}=-\widetilde{\mathbf{c}}^{\left(m,n\right)}_{i_{mn}}\widetilde{\mathbf{B}}^{{\left(m,n\right)}^{\mathrm{T}}}_{i_{mn}} \\
& \widetilde{\mathbf{c}}^{\left(m,n\right)}_{i_{mn}}=\mathbf{c}^{\left(m,n\right)}_{i_{mn}}\mathbf{P}^{\left(m,n\right)}_{i_{mn}},\; \widetilde{\mathbf{B}}^{\left(m,n\right)}_{i_{mn}}=\mathbf{B}^{\left(m,n\right)}\mathbf{P}^{\left(m,n\right)}_{i_{mn}}
\end{aligned}
\right..
\end{equation*}

\begin{figure}
\centering
\includegraphics[scale=0.3]{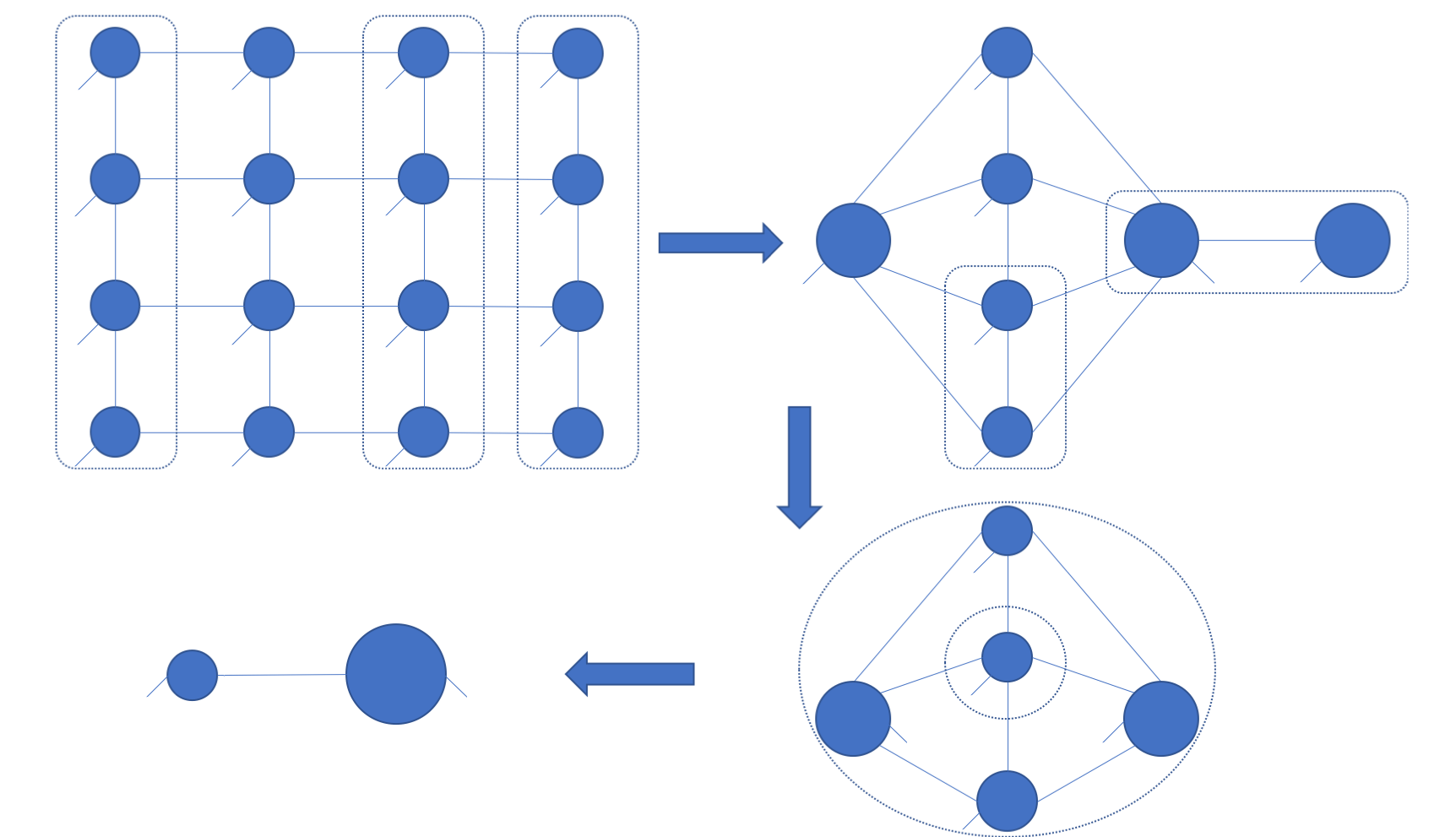}
\quad\quad\quad\quad
\caption{The illustration of the SL method. In the fourth diagram, the small circle means the activated tensor and the large one means the environmental tensor.}
\label{TG contraction}
\end{figure}

To efficiently obtain $\mathbf{B}^{\left(m,n\right)}$, the single layer (SL) method can be used \cite{pivzorn2011time}. As illustrated in Fig. \ref{TG contraction}, the SL first contracts the factors on both left and right sides of the activated tensor, and it contracts the resultant tensors again until this TT is surrounded by a single layer on both sides. It then contracts the remained TT factors except the activated tensor such that the activated tensor is surrounded by four tensors. Finally, the environmental tensor is derived by contracting the four tensors.

The pseudocode of low-rank TG tensor completion is outlined in Algorithm \ref{algorithm_ALS-TG}.
\begin{algorithm}
\caption{Alternating least squares for low-rank tensor grid completion (ALS-TG)}
\label{algorithm_ALS-TG}
\begin{algorithmic}[1]
\REQUIRE Zero-filled tensor $\mathcal{T}$, binary tensor $\mathcal{W}$, the maximal number of iterations $K$
\ENSURE The recovered tensor $\mathcal{X}$, the set of TG factors $\left\{\mathcal{C}\right\}$
\STATE Apply Algorithm \ref{algorithm_2SDMRG} to initialize TG factors $\left\{\mathcal{C}\right\}$
\FOR{$k=1$ \TO $K$}
\FOR{$m=1$ \TO $M$}
\FOR{$n=1$ \TO $N$}
\FOR{$i_{mn}=1$ \TO $I_{mn}$}
\STATE Update $\mathcal{C}^{\left(m,n\right)}_{:,i_{mn}}$ according to (\ref{update})
\ENDFOR
\ENDFOR
\ENDFOR
\STATE $\mathcal{X}=\operatorname{TG}\left(\left\{\mathcal{C}\right\}\right)$
\IF{converged}
\STATE break
\ENDIF
\ENDFOR
\RETURN{$\mathcal{X}$, $\left\{\mathcal{C}\right\}$}
\end{algorithmic}
\end{algorithm}

\subsection{Computational complexity}

In the computational complexity analysis, we assume $\mathcal{X}\in \mathbb{R}^{I_1\times \dotsm \times I_D}$ and $M$ is the number of samples. The computational complexity mainly comes from two parts. First, in the optimization of the $i_d$-th sub-sub-problem, ALS-TG computes the Hessian matrix $\mathbf{H}^{\left(d\right)}_{i_d}$ and its inverse matrix, which costs $O\left(mR^8/I_d\right)$ and $O\left(R^{12}\right)$, respectively. Hence the update of the $d$-th TG factor requires $O\left(mR^8\right)$ and $O\left(I_dR^{12}\right)$, respectively. In one iteration, the computational complexity is $\max\left\{O\left(mDR^{8}\right),O\left(R^{12}\sum^{D}_{d=1}I_d\right)\right\}=O\left(mDR^{8}\right)$. The other part perspective is the computation of $\mathbf{B}^{\left(m,n\right)}$, which costs $O\left(R^{\sqrt{D}}\prod^{D}_{n=1,n\neq d}I_n\right)$. Thus it costs $O\left(R^{\sqrt{D}}\prod^{D}_{d=1}I_d\right)$ in one iteration. 

Therefore, the total computational complexity in one iteration is
\begin{align*}
\max{\left\{ O\left(mDR^8\right),O\left(R^{\sqrt{D}}\prod^{D}_{d=1}I_d\right) \right\}}.
\end{align*}

\section{Parallel Matrix Factorization for Tensor Grid}
\label{section_accelerated_algorithm}

The main computational cost of ALS-TG comes from computing the environmental tensor, which is frequently computed. The computational cost can be greatly reduced if we can avoid the computation of environmental tensor. 

The main idea here is to utilize the parallel matrix factorization for approximating the tensor unfolding. Here we substitute $M+N-2$ pairs of products of factors for the tensor unfoldings. 

\begin{figure}
\centering
\includegraphics[scale=0.4]{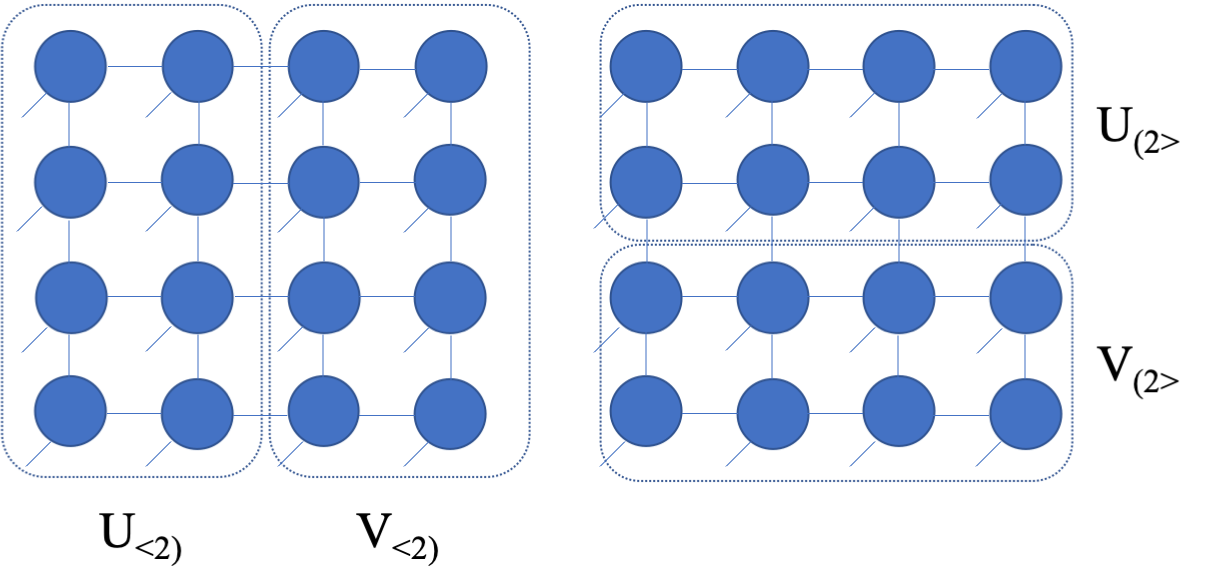}
\caption{The illustration of parallel matrix factorization for a $4\times 4$ TG, where $\mathbf{U}_{<2)}$ and $\mathbf{V}_{<2)}$ are the row factors, and $\mathbf{U}_{(2>}$ and $\mathbf{V}_{(2>}$ are the column factors.}
\label{parallel TG factorization}
\end{figure}

As it shows in Fig. \ref{parallel TG factorization}, there is a TT of length $N$ if we regard $M$ TG factors of $n$-th column as a TT factor $\mathcal{G}^{\left(1,n\right)}$. This perspective gives us $N-1$ pairs of row factors $\mathbf{U}_{<j)},\mathbf{V}_{<j)}$ for TG approximation $\mathbf{U}_{<j)}\mathbf{V}_{<j)}\approx \mathbf{T}_{<j)}$, where $\mathbf{U}_{<j)}\in \mathbb{R}^{\prod^M_{m=1}\prod^j_{n=1}I_{mn}\times \bar{R}_j}$, $\mathbf{V}_{<j)}\in \mathbb{R}^{\bar{R}_j\times \prod^M_{m=1}\prod^N_{n=j+1}I_{mn}}$ and $\mathbf{T}_{<j)}\in \mathbb{R}^{\prod^M_{m=1}\prod^j_{n=1}I_{mn}\times \prod^M_{m=1}\prod^N_{n=j+1}I_{mn}}$. Similarly, there is a TT of length $M$ if we regard $N$ TG factors of $m$-th row as a TT factor $\mathcal{G}^{\left(2,m\right)}$, which gives us $M-1$ pairs of column factors $\mathbf{U}_{(i>},\mathbf{V}_{(i>}$ for TG approximation $\mathbf{U}_{(i>}\mathbf{V}_{(i>}\approx \mathbf{T}_{(i>}$, where $\mathbf{U}_{(i>}\in \mathbb{R}^{\prod^i_{m=1}\prod^N_{n=1}I_{mn}\times \bar{R}_{i+N-1}}$, $\mathbf{V}_{(j>}\in \mathbb{R}^{\bar{R}_{i+N-1}\times \prod^M_{m=i+1}\prod^N_{n=1}I_{mn}}$ and $\mathbf{T}_{(i>}\in \mathbb{R}^{\prod^i_{m=1}\prod^N_{n=1}I_{mn}\times \prod^M_{m=i+1}\prod^N_{n=1}I_{mn}}$. The rank parameters $\left[ \bar{R}_1,\dotsc,\bar{R}_{M+N-2} \right]$ need to be pre-defined.

Introducing the auxiliary variables $\mathcal{M}_i$, $i=1\dotsc,M-1$ and $\mathcal{M}_j$, $j=1\dotsc,N-1$, we formulate the corresponding model as follows:
\begin{align}
\min_{\substack{\mathbf{U}_{(i>}, \mathbf{V}_{(i>} \\ \mathbf{U}_{<j)}, \mathbf{V}_{<j)} \\ i=1\dotsc,M-1 \\ j=1\dotsc,N-1}}\;\;& \frac{1}{2}\sum^{N-1}_{j=1}w_i\lVert \mathbf{U}_{<j)}\mathbf{V}_{<j)}-\mathbf{M}_{<j)} \rVert^2_{\mathrm{F}}+ \notag \\
& \frac{1}{2}\sum^{M-1}_{i=1}w_{i+N-1}\lVert \mathbf{U}_{(i>}\mathbf{V}_{(i>}-\mathbf{M}_{(i>} \rVert^2_{\mathrm{F}} \notag \\
\text{s. t.\quad\;}\;\;& \operatorname{P}_{\mathbb{O}}\left(\mathcal{X}\right)=\operatorname{P}_{\mathbb{O}}\left(\mathcal{T}\right).
\label{TG completion model2}
\end{align}
This method is called parallel matrix factorization for TG completion (PMac-TG). Aside from its less computational complexity, another advantage of this method is the utilization of well-balanced matricization scheme, i.e., the unfolding is balanced (square), which is able to capture more hidden information and hence performs better \cite{mu2014square, bengua2017efficient, huang2019provable}.

\subsection{Algorithm}

To solve the problem (\ref{TG completion model2}), we alternately optimize the row factors and column factors. This is a simple linear fitting problem, the updating scheme contains \cite{xu2013parallel, tan2014tensor}:
\begin{align}
\mathbf{U}^k_{<j)}=\mathbf{M}^{k-1}_{<j)}\mathbf{V}^{k-1^{\mathrm{T}}}_{<j)},\; \mathbf{U}^k_{(i>}=\mathbf{M}^{k-1}_{(i>}\mathbf{V}^{k-1^{\mathrm{T}}}_{(i>},
\label{updateU}
\end{align}
\begin{align}
& \mathbf{V}^k_{<j)}=\left(\mathbf{U}^{k^{\mathrm{T}}}_{<j)}\mathbf{U}^k_{<j)}\right)^{\dagger}\mathbf{U}^{k^{\mathrm{T}}}_{<j)}\mathbf{M}^{k-1}_{<j)}, \notag \\ 
& \mathbf{V}^k_{(i>}= \left(\mathbf{U}^{k^{\mathrm{T}}}_{(i>}\mathbf{U}^k_{(i>}\right)^{\dagger}\mathbf{U}^{k^{\mathrm{T}}}_{(i>}\mathbf{M}^{k-1}_{(i>}
\label{updateV}
\end{align}
and 
\begin{align}
\mathbf{M}^k_{<j)}=\mathbf{U}^k_{<j)}\mathbf{V}^k_{<j)},\; \mathbf{M}^k_{(i>}=\mathbf{U}^k_{(i>}\mathbf{V}^k_{(i>},
\label{updateM}
\end{align}
where the superscript $k$ denotes the current iteration. As one of its advantages, this updating scheme avoids computing $\left(\mathbf{V}^k_{(i>}\mathbf{V}^{k^{\mathrm{T}}}_{(i>}\right)^{\dagger}$ and $\left(\mathbf{V}^k_{<j)}\mathbf{V}^{k^{\mathrm{T}}}_{<j)}\right)^{\dagger}$ since $\mathbf{U}^k_{<j)}\mathbf{V}^k_{<j)}$ and $\mathbf{U}^k_{(i>}\mathbf{V}^k_{(i>}$ suffice to generate the recovered tensor.

The recovered tensor can be updated as follows:
\begin{align}
\mathcal{X}^k_{\mathbb{O}}=\mathcal{T}_{\mathbb{O}},\; \mathcal{X}^k_{\mathbb{O}^{\perp}}=\operatorname{P}_{\mathbb{O}^{\perp}}\left( \sum^{N-1}_{j=1}w_j\mathcal{M}^k_j+\sum^{M-1}_{i=1}w_{i+N-1}\mathcal{M}^k_i \right).
\label{updateX}
\end{align}

The pseudocode of PMac-TG is given in Algorithm \ref{algorithm_PMac-TG}. 
\begin{algorithm}
\caption{Tensor completion by parallel matrix factorization via tensor grid (PMac-TG)}
\label{algorithm_PMac-TG}
\begin{algorithmic}[1]
\REQUIRE Zero-filled tensor $\mathcal{T}$, binary tensor $\mathcal{W}$, initial ranks of parallel factors, initial weight vector $\mathbf{w}$, the maximal \# iterations $K$
\ENSURE The recovered tensor $\mathcal{X}$
\STATE Initialize $\left\{ \mathbf{U}^0,\mathbf{V}^0 \right\}$ and $\mathcal{X}^0=\mathcal{T}$
\FOR{$k=1$ \TO $K$}
\FOR{$j=1$ \TO $N-1$}
\STATE update the left row factor according to (\ref{updateU})
\STATE update the right row factors according to (\ref{updateV})
\STATE update the auxiliary tensor according to (\ref{updateM})
\ENDFOR
\FOR{$i=1$ \TO $M-1$}
\STATE update the left column factor according to (\ref{updateU})
\STATE update the right column factor according to (\ref{updateV})
\STATE update the auxiliary tensor according to (\ref{updateM})
\ENDFOR
\STATE update the recovered tensor according to (\ref{updateX})
\IF{converged}
\STATE break
\ENDIF
\ENDFOR
\RETURN{$\mathcal{X}$}
\end{algorithmic}
\end{algorithm}

\subsection{Computational Complexity}

Without loss of generality, we assume $\mathcal{X}\in \mathbb{R}^{I_1\times \dotsm \times I_D}$. The main complexity results from two parts. The computation of $\mathbf{U}^{k^{\mathrm{T}}}_{<i)}\mathbf{U}^k_{<i)}$ costs $O\left(\sqrt{D}R\prod^{D}_{d=1}I_d\right)$. The computation of the inverse of $\mathbf{U}^{k^{\mathrm{T}}}_{<i)}\mathbf{U}^k_{<i)}$ costs $O\left(R^{3\sqrt{D}}\right)$. Therefore, the total computational cost in one iteration of PMac-TG is $\max\left\{ O\left(\sqrt{D}R\prod^{D}_{d=1}I_d\right),O\left(R^{3\sqrt{D}}\right) \right\}=O\left(\sqrt{D}R\prod^{D}_{d=1}I_d\right)$, which is much smaller than that of ALS-TG.

\section{Numerical Experiments}
\label{section_experiment}
 
In this section, the proposed algorithms are tested on synthetic data and real-world data. Seven algorithms are benchmarked in the experiments, including low-rank tensor ring completion via alternating least square (TR-ALS) \cite{wang2017efficient}, low-rank tensor train completion via parallel matrix factorization (TMac-TT) \cite{bengua2017efficient}, low-rank tensor train completion via alternating least square (TT-ALS) \cite{wang2016tensor}, high accuracy low-rank tensor completion (HaLRTC) \cite{liu2013tensor}, non-negative CP completion via alternating proximal gradient (CP-APG) \cite{xu2013block} and the proposed two methods, namely ALS-TG and PMac-TG.

In Algorithm \ref{algorithm_PMac-TG}, the weights $w_k$, $k=1,\dotsc,M+N-2$ are chosen as follows:
\begin{equation}
\label{weight}
\left\{
\begin{aligned}
& w_k=\min \left\{\prod^{k}_{j=1}\prod^{M}_{m=1}I_{mj},\prod^{N}_{j=k+1}\prod^{M}_{m=1}I_{mj}\right\}, \\
& k=1,\dotsc,N-1 \\
& w_{k+N-1}=\min \left\{\prod^{k}_{j=1}\prod^{N}_{n=1}I_{jn},\prod^{M}_{j=k+1}\prod^{N}_{n=1}I_{jn}\right\}, \\
& k=1,\dotsc,M-1
\end{aligned}
\right..
\end{equation}
We determine the initial ranks $\left[ \bar{R}_1,\dotsc,\bar{R}_{M+N-2} \right]$ as follows:
\begin{equation}
\left\{
\begin{aligned}
& \bar{R}_i=\min\left\{ i_0|\frac{\sigma^{<i)}_{i_0}}{\sigma^{<i)}_1}<th,\; i=1,\dotsc,N-1 \right\} \\
& \bar{R}_{j+N-1}=\min\left\{ j_0|\frac{\sigma^{(j>}_{j_0}}{\sigma^{(j>}_1}<th,\; j=1,\dotsc,M-1 \right\}
\end{aligned}
\right.,
\end{equation}
where $\sigma^{<j)}_1$ and $\sigma^{<j)}_{j_0}$ are the largest and the $j_0$-th singular values of $\mathbf{M}_{<j)}$, $\sigma^{(i>}_1$ and $\sigma^{(i>}_{i_0}$ are the largest and the $i_0$-th singular values of $\mathbf{M}_{(i>}$, provided that the singular values are on the descent order. We set thresholding $th=0.02$ empirically. Alternatively, we can initialize the ranks by setting $\bar{R}_1=\dotsm=\bar{R}_{M+N-2}=\bar{R}_0$, where $\bar{R}_0$ is a pre-defined number. To distinguish the two approaches, we refer to the manual way as PMac-TG, which is PMac-TG with manual rank determination, and refer to the automatic way as A-PMac-TG, which is PMac-TG with automatic rank determination.

In tensor sampling, we use sampling rate (SR) to denote the ratio of the number of the samples to the number the tensor entries.

Three kinds of metrics are used to quantitatively measure the recovery performance. The first one is relative error (RE), which is defined as $\mathrm{RE}=\lVert \hat{\mathcal{X}}-\mathcal{X}_0 \rVert_{\mathrm{F}}/\lVert \mathcal{X}_0 \rVert_{\mathrm{F}}$, where $\mathcal{X}_0$ is the ground truth and $ \hat{\mathcal{X}} $ is the estimate. The second one is peak signal-to-noise ratio (PSNR), and the third one is structural similarity index (SSIM), which are commonly used evaluations for image quality.

Two stopping rules are used in the algorithms. The relative change (RC) is defined as $\mathrm{RC}=\lVert \mathcal{X}^k-\mathcal{X}^{k-1} \rVert_{\mathrm{F}}/\lVert \mathcal{X}^{k-1} \rVert_{\mathrm{F}}$, where the superscript $k$ means the current iteration. The algorithm terminates when $\text{RC}<1\times 10^{-8}$ or it reaches the maximal iteration $K$. To be fair, we fix $K=100$ for all algorithms in the experiments.

In the following parts, we use one parameter to denote the tensor rank for simplicity if the rank is a vector with all elements equal. For example, we mean the TG rank is $\left[2,\dotsc,2\right]$ by saying the TG rank is $2$.

All the experiments are conducted in MATLAB 9.3.0 on a computer with a 2.8GHz CPU of Intel Core i7 and a 16GB RAM.

\subsection{Synthetic data}

In this subsection, we perform two experiments on randomly generated tensors. The rank is known in advance. A successful recovery indicates $\text{RE}< 1\times 10^{-6}$.

\begin{figure*}[htbp]
\centering
\begin{subfigure}[t]{0.45\textwidth}
\centering
\includegraphics[scale=0.22]{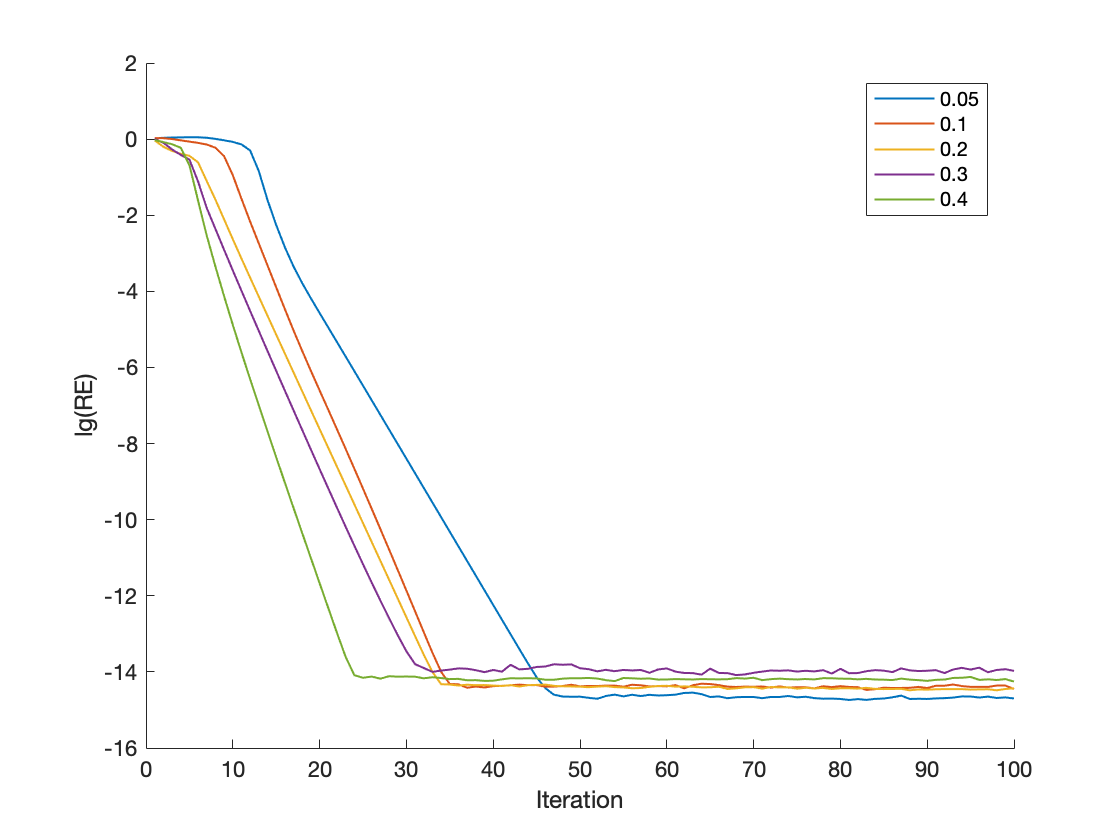}
\subcaption{Convergence results of ALS-TG with sampling rates varying from $0.05$ to $0.5$ for a $9$-order tensor of size $4\times \dotsm \times 4$ and TG rank $2$.}
\end{subfigure}
\qquad
\begin{subfigure}[t]{0.45\textwidth}
\centering
\includegraphics[scale=0.22]{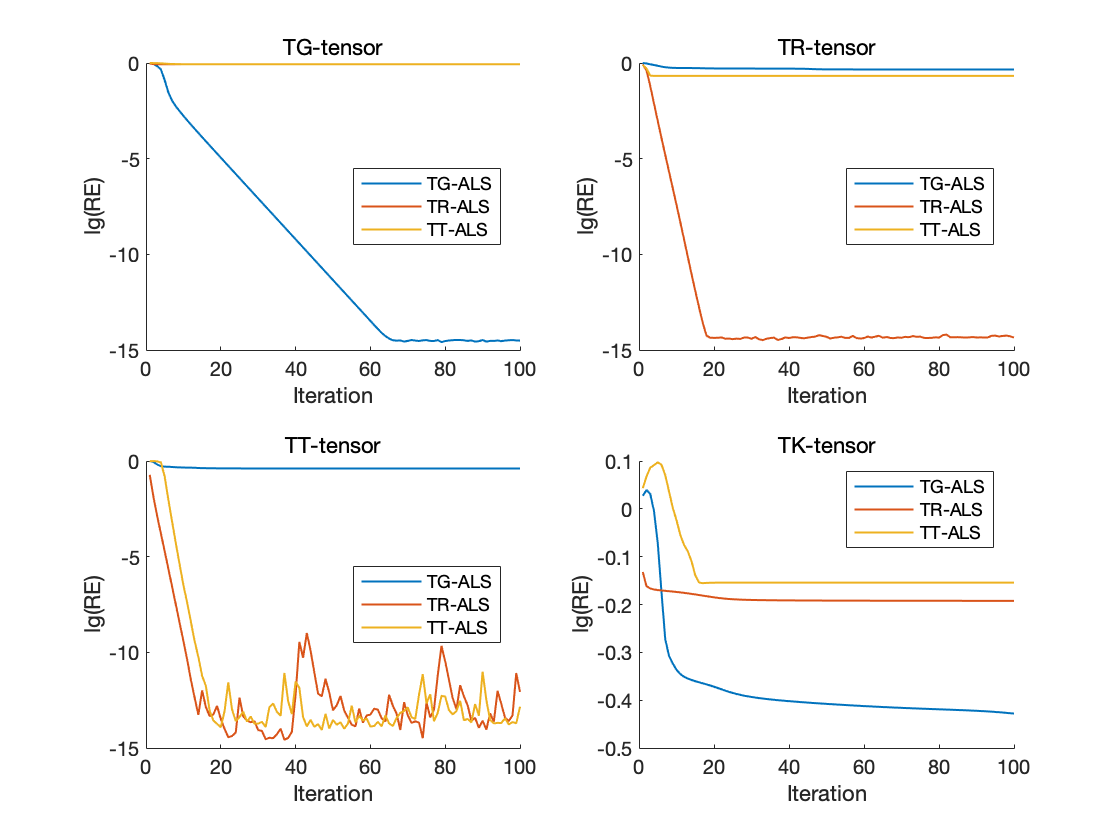}
\subcaption{The cross completion of the TG-tensor, the TR-tensor, the TT-tensor and the TK-tensor performed by ALS-TG, TR-ALS and TT-ALS.}
\end{subfigure}
\caption{The results of synthetic data completion for $9$-order tensors of size $4\times \dotsm \times 4$. The TG rank, TR rank, TT rank and TK rank are all $2$.}
\label{result_synthetic1-2}
\end{figure*}

We first consider a $9$-order tensor $\mathcal{X}\in \mathbb{R}^{4\times \dotsm \times 4}$ of TG rank $2$. The tensor is generated by TG contraction, and each entry of TG factors obeys the standard normal distribution $\operatorname{N}\left(0,1\right)$. Fig. \ref{result_synthetic1-2}(a) shows the convergence of ALS-TG. The sampling rates are $0.05$, $0.1$, $0.2$, $0.3$, $0.4$ and $0.5$. The results show the algorithm converges faster with increasing sampling rate. But it is unable to recover the tensor if the sampling rate is less than $0.05$. The convergence demonstrates the effectiveness of ALS-TG.

We then consider four synthetic tensors that are generated by TG contraction, TR contraction, TT contraction and TK contraction. Every entry of each factor is sampled from the standard normal distribution $\operatorname{N}\left(0,1\right)$. The TG rank, TR rank, TT rank and TK rank are all $2$, and the sampling rates are all $0.1$. We apply ALS-TG, TR-ALS and TT-ALS to recover the four tensors. Fig. \ref{result_synthetic1-2}(b) shows the results of relative errors versus iteration numbers for the three algorithms.

\begin{figure}
\centering
\includegraphics[scale=0.24]{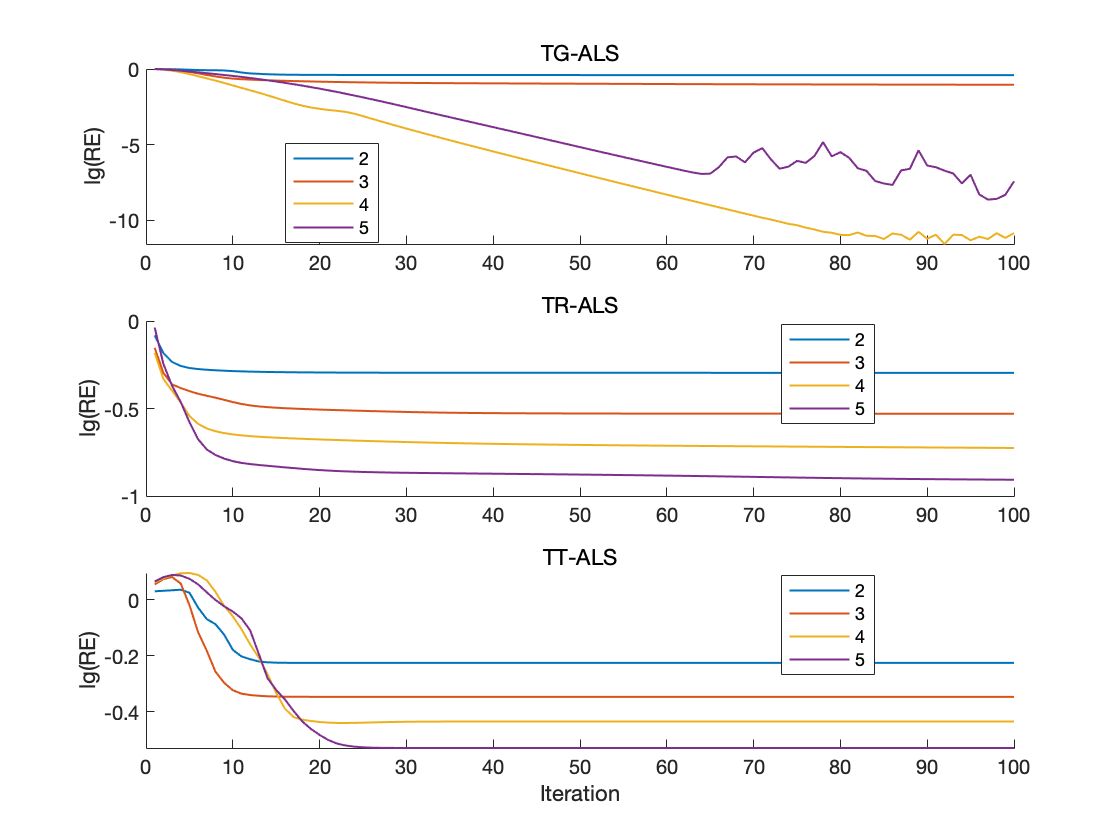}
\caption{The completion results of a $9$-order TK-tensor of size $4\times \dotsm \times 4$ with TK rank $2$, performed by ALS-TG, TR-ALS and TT-ALS with rank being all $2$, $3$, $4$ and $5$.}
\label{result_synthetic3}
\end{figure}

The results illustrate that each algorithm successfully recovers the tensor generated by their corresponding decompositions, i.e., ALS-TG recovers the TG-tensor and TR-ALS recovers the TR-tensor, etc. However, TR-ALS can also recover the TT-tensor, but TT-ALS fails to recover the TR-tensor. Though all three methods are unable to recover the TK-tensor, ALS-TG shows better capability of recovering the TK-tensor than TR-ALS and TT-ALS. Fig. \ref{result_synthetic3} shows the recovery results of TK-tensor by the three methods. The TG rank, TR rank and TT rank vary from $2$ to $5$. The results show that only ALS-TG can recover the TK-tensor with TG rank $4$, which validates that ALS-TG outperforms TR-ALS and TT-ALS in low TK rank tensor completion.

\subsection{Color images}

In this section, we first consider the completion of the RGB image \emph{lena} of size $256\times 256\times 3$. To enhance the performance of completion, we use a tensorization operation \cite{bengua2017efficient} that contains reshaping and reordering operations. This method augments the original tensor into a higher-order tensor, permutes its order, and reshapes the resulting tensor into the third one. For example, we reshape a low-order tensor of size $M_1\dotsm M_L\times N_1\dotsm N_L\times 3$ into a tensor of size $M_1\times M_2 \times \dotsm \times M_L\times N_1\times N_2 \times \dotsm \times N_L\times 3$,  reorder and reshape it to obtain a tensor of size $M_1N_1\times M_2N_2 \times \dotsm \times M_LN_L\times 3$. Here we choose $M_l=N_l=2$, $l=1,\dotsc,8$ for \emph{lena}. The sampling rates are $0.1$, $0.2$ and $0.3$. The size of TG is $3\times 3$.

\begin{figure}[htbp]
\centering
\includegraphics[scale=0.22]{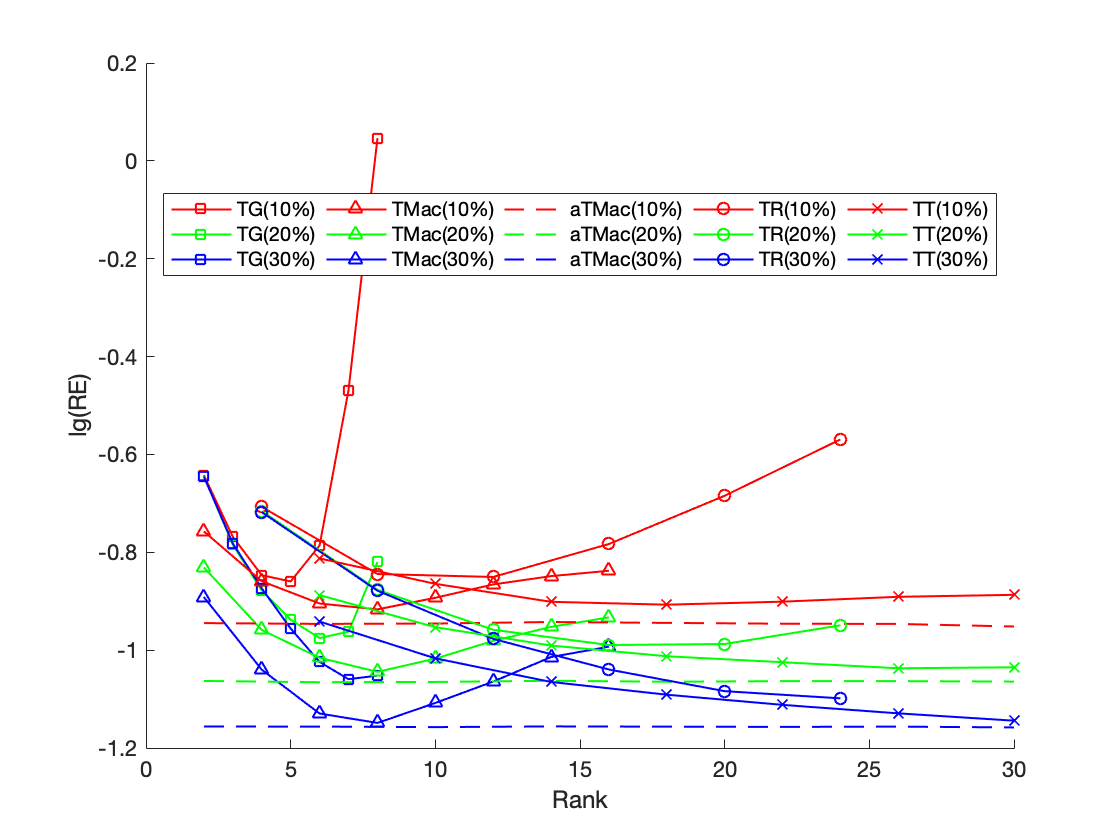}
\caption{The completion results of \emph{lena} based on ALS-TG, PMac-TG, A-PMac-TG, TR-ALS and TMac-TT algorithms. The sampling rates are $0.1$, $0.2$ and $0.3$. The rank ranges from $2$ to $30$.}
\label{result_image1}
\end{figure}

Fig. \ref{result_image1} shows the relative error versus tensor rank with different sampling rates. The ``TG'' label represents the results derived by ALS-TG algorithm, ``PMac'' label represents the results derived by PMac-TG algorithm and so on. As we can see, when the sampling rate increases, the best rank also becomes large. The reason is the real-world data is not strictly low-rank and the tensor rank that leads to well performance becomes large with the increasing number of the observations. On the other hand, the relative error firstly decreases as the rank increases from a small value to the best rank, and then starts to increase once the rank is greater than the best rank. This phenomenon indicates that with rank enlarged, an under-fitting problem happens first, then an over-fitting problem occurs, where the over-fitting problem is also discovered in \cite{wang2017efficient}. The result also shows the best TG rank, TR rank and TT rank gradually increase under the same setting of sampling rate. This is a validation of the more powerful representation ability of TG structure. The PMac-TG and A-PMac-TG methods show the lowest relative errors with its best ranks among all sampling rates.

The other group of experiments is to recover the RGB image \emph{house} which is masked by texts. The initial ranks of all algorithms are well-tuned. The best TG rank, TG factor rank, TR rank, TT rank, CP rank are $5$, $18$, $10$, $18$ and $20$, respectively. The HaLRTC and LRMC algorithms does not require the initial rank since they minimize the rank. We use RE and SSIM for evaluation.

\begin{figure*}[htbp]
\centering
\begin{subfigure}[t]{0.15\textwidth}
\centering
\setlength{\abovecaptionskip}{0pt}
\setlength{\belowcaptionskip}{-2pt}
\subcaption*{Original}
\includegraphics[width=1\textwidth]{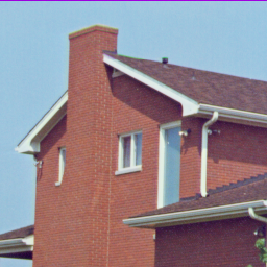}
\subcaption*{RE, SSIM}
\end{subfigure}
\begin{subfigure}[t]{0.15\textwidth}
\centering
\setlength{\abovecaptionskip}{0pt}
\setlength{\belowcaptionskip}{-2pt}
\subcaption*{ALS-TG, 503.167s}
\includegraphics[width=1\textwidth]{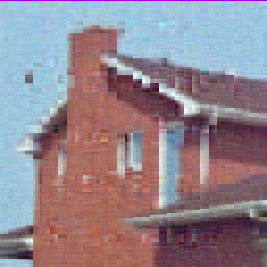}
\subcaption*{0.111, 0.8513}
\end{subfigure}
\begin{subfigure}[t]{0.15\textwidth}
\centering
\setlength{\abovecaptionskip}{0pt}
\setlength{\belowcaptionskip}{-2pt}
\subcaption*{A-PMac-TG, 2.64s}
\includegraphics[width=1\textwidth]{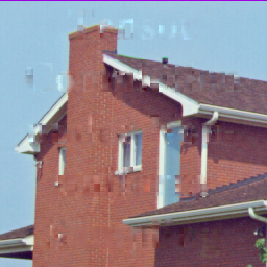}
\subcaption*{0.070, 0.9559}
\end{subfigure}
\begin{subfigure}[t]{0.15\textwidth}
\centering
\setlength{\abovecaptionskip}{0pt}
\setlength{\belowcaptionskip}{-2pt}
\subcaption*{TMac-TT, 6.25s}
\includegraphics[width=1\textwidth]{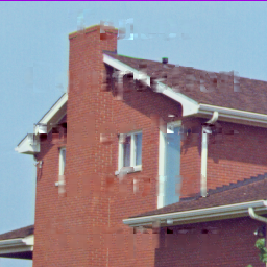}
\subcaption*{0.075, 0.9577}
\end{subfigure}
\begin{subfigure}[t]{0.15\textwidth}
\centering
\setlength{\abovecaptionskip}{0pt}
\setlength{\belowcaptionskip}{-2pt}
\subcaption*{CP-APG, \textbf{1.92s}}
\includegraphics[width=1\textwidth]{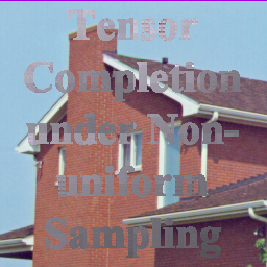}
\subcaption*{0.131, 0.8254}
\end{subfigure}

\begin{subfigure}[t]{0.15\textwidth}
\centering
\setlength{\abovecaptionskip}{0pt}
\setlength{\belowcaptionskip}{-2pt}
\subcaption*{Observed}
\includegraphics[width=1\textwidth]{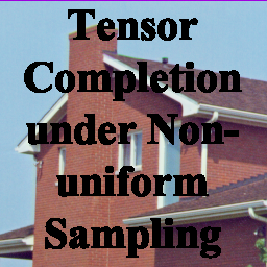}
\subcaption*{RE, SSIM}
\end{subfigure}
\begin{subfigure}[t]{0.15\textwidth}
\centering
\setlength{\abovecaptionskip}{0pt}
\setlength{\belowcaptionskip}{-2pt}
\subcaption*{PMac-TG, 2,49s}
\includegraphics[width=1\textwidth]{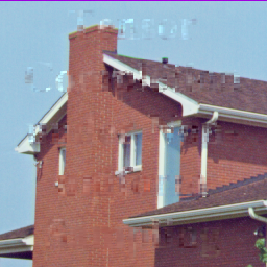}
\subcaption*{\textbf{0.062}, \textbf{0.9639}}
\end{subfigure}
\begin{subfigure}[t]{0.15\textwidth}
\centering
\setlength{\abovecaptionskip}{0pt}
\setlength{\belowcaptionskip}{-2pt}
\subcaption*{TR-ALS, 159.95s}
\includegraphics[width=1\textwidth]{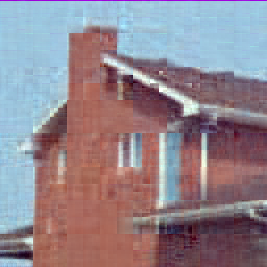}
\subcaption*{0.098, 0.8706}
\end{subfigure}
\begin{subfigure}[t]{0.15\textwidth}
\centering
\setlength{\abovecaptionskip}{0pt}
\setlength{\belowcaptionskip}{-2pt}
\subcaption*{HaLRTC, 7.94s}
\includegraphics[width=1\textwidth]{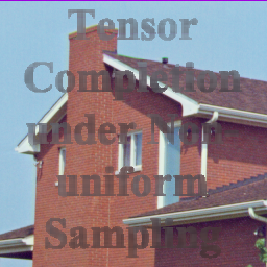}
\subcaption*{0.126, 0.8978}
\end{subfigure}
\begin{subfigure}[t]{0.15\textwidth}
\centering
\setlength{\abovecaptionskip}{0pt}
\setlength{\belowcaptionskip}{-2pt}
\subcaption*{LRMC, 6.78s}
\includegraphics[width=1\textwidth]{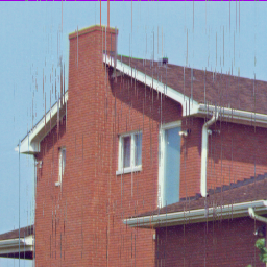}
\subcaption*{0.065, 0.9624}
\end{subfigure}
\caption{The recovery results of \emph{house} under a non-uniform sampling based on various algorithms, including ALS-TG, PMac-TG, A-PMac-TG, TR-ALS, TMac-TT, HaLRTC, CP-APG and LRMC. The best TG rank, TG factor rank, TR rank, TT rank, CP rank are $5$, $18$, $10$, $18$ and $20$, respectively.}
\label{result_image2}
\end{figure*}

Fig. \ref{result_image2} displays the recovery results. The PMac-TG and A-PMac-TG methods give relative errors around $0.06$ and SSIMs above $0.96$ within $3$ seconds. In contrast, the ALS-TG and TR-ALS algorithms give relative errors around $0.10$ with more than $100$ seconds. The HaLRTC and CP-APG give relative errors around $0.12$. However, their SSIMs are lower than those of the former ones. The LRMC algorithm shows the lower relative error and higher SSIM than these of all methods except PMac-TG. The result shows our method is very robust for non-uniform sampling.

\subsection{Hyperspectral image}

In this subsection we benchmark these methods on a hyperspectral image named ``jasper ridge'' \cite{ zhu2014spectral, zhu2014structured}. There are $512\times 614$ pixels in it and each pixel is recorded at $198$ channels ranging from $380$ nm to $2500$ nm. We consider a sub-image of $100\times 100$ pixels where the first pixel starts from the $\left(105,269\right)$-th pixel in the original image. Accordingly, we obtain a tensor of size $100\times 100\times 3\times 198$, further reshape it into a tensor of size $10\times 10\times 10\times 10\times 11\times 18$. We randomly sample $10\%$ pixels of this video. The size of TG is $3\times 2$.

\begin{figure*}[htbp]
\centering
\begin{subfigure}[t]{0.15\textwidth}
\centering
\setlength{\abovecaptionskip}{0pt}
\setlength{\belowcaptionskip}{0pt}
\subcaption*{Original}
\includegraphics[width=1\textwidth]{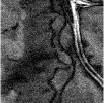}
\end{subfigure}
\begin{subfigure}[t]{0.15\textwidth}
\centering
\setlength{\abovecaptionskip}{0pt}
\setlength{\belowcaptionskip}{0pt}
\subcaption*{Observed}
\includegraphics[width=1\textwidth]{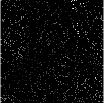}
\end{subfigure}

\begin{subfigure}[t]{0.15\textwidth}
\centering
\setlength{\abovecaptionskip}{0pt}
\setlength{\belowcaptionskip}{0pt}
\subcaption*{TG(3)}
\includegraphics[width=1\textwidth]{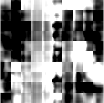}
\end{subfigure}
\begin{subfigure}[t]{0.15\textwidth}
\centering
\setlength{\abovecaptionskip}{0pt}
\setlength{\belowcaptionskip}{0pt}
\subcaption*{PMac-TG(30)}
\includegraphics[width=1\textwidth]{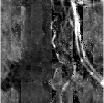}
\end{subfigure}
\begin{subfigure}[t]{0.15\textwidth}
\centering
\setlength{\abovecaptionskip}{0pt}
\setlength{\belowcaptionskip}{0pt}
\subcaption*{TR(5)}
\includegraphics[width=1\textwidth]{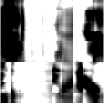}
\end{subfigure}
\begin{subfigure}[t]{0.15\textwidth}
\centering
\setlength{\abovecaptionskip}{0pt}
\setlength{\belowcaptionskip}{0pt}
\subcaption*{TMac-TT(30)}
\includegraphics[width=1\textwidth]{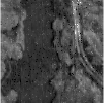}
\end{subfigure}
\begin{subfigure}[t]{0.15\textwidth}
\centering
\setlength{\abovecaptionskip}{0pt}
\setlength{\belowcaptionskip}{0pt}
\subcaption*{CP-APG(10)}
\includegraphics[width=1\textwidth]{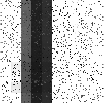}
\end{subfigure}
\begin{subfigure}[t]{0.15\textwidth}
\centering
\setlength{\abovecaptionskip}{0pt}
\setlength{\belowcaptionskip}{0pt}
\subcaption*{A-PMac-TG}
\includegraphics[width=1\textwidth]{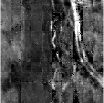}
\end{subfigure}

\begin{subfigure}[t]{0.15\textwidth}
\centering
\setlength{\abovecaptionskip}{0pt}
\setlength{\belowcaptionskip}{0pt}
\subcaption*{TG(3)}
\includegraphics[width=1\textwidth]{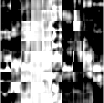}
\end{subfigure}
\begin{subfigure}[t]{0.15\textwidth}
\centering
\setlength{\abovecaptionskip}{0pt}
\setlength{\belowcaptionskip}{0pt}
\subcaption*{PMac-TG(30)}
\includegraphics[width=1\textwidth]{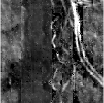}
\end{subfigure}
\begin{subfigure}[t]{0.15\textwidth}
\centering
\setlength{\abovecaptionskip}{0pt}
\setlength{\belowcaptionskip}{0pt}
\subcaption*{TR(5)}
\includegraphics[width=1\textwidth]{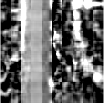}
\end{subfigure}
\begin{subfigure}[t]{0.15\textwidth}
\centering
\setlength{\abovecaptionskip}{0pt}
\setlength{\belowcaptionskip}{0pt}
\subcaption*{TMac-TT(30)}
\includegraphics[width=1\textwidth]{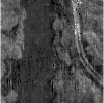}
\end{subfigure}
\begin{subfigure}[t]{0.15\textwidth}
\centering
\setlength{\abovecaptionskip}{0pt}
\setlength{\belowcaptionskip}{0pt}
\subcaption*{CP-APG(10)}
\includegraphics[width=1\textwidth]{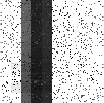}
\end{subfigure}
\begin{subfigure}[t]{0.15\textwidth}
\centering
\setlength{\abovecaptionskip}{0pt}
\setlength{\belowcaptionskip}{0pt}
\subcaption*{HaLRTC}
\includegraphics[width=1\textwidth]{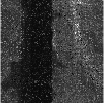}
\end{subfigure}

\begin{subfigure}[t]{0.15\textwidth}
\centering
\setlength{\abovecaptionskip}{0pt}
\setlength{\belowcaptionskip}{0pt}
\subcaption*{TG(3)}
\includegraphics[width=1\textwidth]{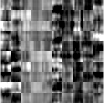}
\end{subfigure}
\begin{subfigure}[t]{0.15\textwidth}
\centering
\setlength{\abovecaptionskip}{0pt}
\setlength{\belowcaptionskip}{0pt}
\subcaption*{PMac-TG(30)}
\includegraphics[width=1\textwidth]{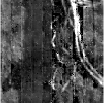}
\end{subfigure}
\begin{subfigure}[t]{0.15\textwidth}
\centering
\setlength{\abovecaptionskip}{0pt}
\setlength{\belowcaptionskip}{0pt}
\subcaption*{TR(5)}
\includegraphics[width=1\textwidth]{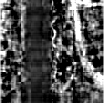}
\end{subfigure}
\begin{subfigure}[t]{0.15\textwidth}
\centering
\setlength{\abovecaptionskip}{0pt}
\setlength{\belowcaptionskip}{0pt}
\subcaption*{TMac-TT(30)}
\includegraphics[width=1\textwidth]{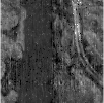}
\end{subfigure}
\begin{subfigure}[t]{0.15\textwidth}
\centering
\setlength{\abovecaptionskip}{0pt}
\setlength{\belowcaptionskip}{0pt}
\subcaption*{CP-APG(10)}
\includegraphics[width=1\textwidth]{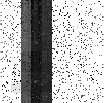}
\end{subfigure}
\begin{subfigure}[t]{0.15\textwidth}
\centering
\setlength{\abovecaptionskip}{0pt}
\setlength{\belowcaptionskip}{0pt}
\subcaption*{LRMC}
\includegraphics[width=1\textwidth]{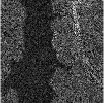}
\end{subfigure}
\caption{The completion results of \emph{jasper ridge} based on various algorithms, including ALS-TG, PMac-TG, A-PMac-TG, TR-ALS, TMac-TT, HaLRTC, CP-APG and LRMC. The TG ranks are $3$, $4$ and $5$. The TG factor ranks are $30$, $40$ and $50$. The TR ranks are $4$, $8$ and $12$. The TT ranks are $30$, $40$ and $50$. The CP ranks are $10$, $20$ and $30$.}
\label{result_hsi1_figure}
\end{figure*}

Fig. \ref{result_hsi1_figure} shows the completion results of the first frame and Table \ref{result_hsi1_table} shows the detailed evaluation of the completion. The PMac-TG achieves the smallest relative error $0.051$ and the highest PSNR $21.63$ dB among the recoveries. The A-PMac-TG does not perform as well as the PMac-TG, which indicates it may suffer from an under-fitting problem. The TR-ALS algorithm is time-consuming. The HaLRTC and CP-APG fail to recover due to their higher relative errors and lower PSNRs.

\begin{table}
\centering
\caption{Recovery performance of \emph{jasper ridge}, where $10\%$ entries are observed.}
\label{result_hsi1_table}
\begin{tabular}{ccccc}
\toprule
Method & Rank & RE & PSNR (dB) & CPU time (s) \\
\hline
\multirow{3}{*}{ALS-TG} & 4 & 0.160 & 15.04 & $1.68\times 10^2$ \\
 & 5 & 0.129 & 15.80 & $3.11\times 10^2$ \\
 & 6 & 0.103 & 17.01 & $4.79\times 10^2$ \\
\hline
\multirow{3}{*}{PMac-TG} & 30 & 0.053 & 21.60 & $2.50\times 10^1$ \\
 & 40 & \textbf{0.051} & \textbf{21.63} & $2.63\times 10^1$ \\
 & 50 & 0.067 & 20.63 & $2.88\times 10^1$ \\
\hline
A-PMac-TG & - & 0.075 & 19.05 & $1.42\times 10^1$ \\
\hline
\multirow{3}{*}{TR-ALS} & 4 & 0.215 & 13.83 & $2.13\times 10^2$ \\
 & 8 & 0.123 & 16.36 & $5.18\times 10^2$ \\
 & 12 & 0.082 & 18.06 & $1.82\times 10^3$ \\
\hline
\multirow{3}{*}{TMac-TT} & 30 & 0.062 & 20.61 & $3.66\times 10^1$ \\
 & 40 & 0.079 & 19.84 & $4.36\times 10^1$ \\
 & 50 & 0.104 & 18.89 & $5.28\times 10^1$ \\
\hline
HaLRTC & - & 0.836 & 8.58 & $6.32\times 10^1$ \\
\hline
\multirow{3}{*}{CP-APG} & 10 & 0.949 & 1.45 & $\textbf{1.13}\times \textbf{10}^{\textbf{0}}$ \\
& 20 & 0.351 & 11.92 & $7.02\times 10^1$ \\
& 30 & 0.351 & 11.93 & $1.16\times 10^2$ \\
\hline
LRMC & - & 0.806 & 10.10 & $4.89\times 10^1$ \\
\bottomrule
\end{tabular}
\end{table}

\subsection{Real-world monitoring videos}

In this subsection, we fist test these methods on a video dataset called \emph{explosion}, which is from high speed camera for explosion \footnote{https://pixabay.com/videos/explosion-inlay-explode-fire-16642/}. This video contains $241$ frames and each frame is an RGB image of size $360 \times 640 \times 3$. We keep its first $81$ frames, and uniformly down-sample each frame to size $90 \times 160 \times 3$. Accordingly, we obtain a tensor of size $90 \times 160 \times 3 \times 81$, further reshape it into a tensor of size $2\times 3\times 3\times 5\times 5\times 4\times 4\times 2\times 3\times 81$, and permute it with order $\left[1, 5, 2, 6, 3, 7, 4, 8, 9, 10\right]$, finally we obtain a tensor of size $10 \times 12 \times 12 \times 10 \times 3 \times 3 \times 3 \times 3 \times 3$. We randomly sample $10\%$ pixels of this video. The size of TG is $3\times 3$.

\begin{figure*}[htbp]
\centering
\begin{subfigure}[t]{0.15\textwidth}
\centering
\setlength{\abovecaptionskip}{0pt}
\setlength{\belowcaptionskip}{0pt}
\subcaption*{Original}
\includegraphics[scale=1]{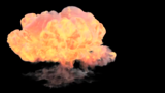}
\end{subfigure}
\begin{subfigure}[t]{0.15\textwidth}
\centering
\setlength{\abovecaptionskip}{0pt}
\setlength{\belowcaptionskip}{0pt}
\subcaption*{Observed}
\includegraphics[scale=1]{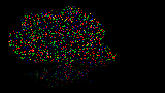}
\end{subfigure}

\begin{subfigure}[t]{0.15\textwidth}
\centering
\setlength{\abovecaptionskip}{0pt}
\setlength{\belowcaptionskip}{0pt}
\subcaption*{TG(3)}
\includegraphics[scale=1]{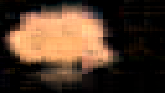}
\end{subfigure}
\begin{subfigure}[t]{0.15\textwidth}
\centering
\setlength{\abovecaptionskip}{0pt}
\setlength{\belowcaptionskip}{0pt}
\subcaption*{PMac-TG(30)}
\includegraphics[scale=1]{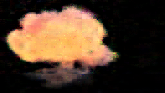}
\end{subfigure}
\begin{subfigure}[t]{0.15\textwidth}
\centering
\setlength{\abovecaptionskip}{0pt}
\setlength{\belowcaptionskip}{0pt}
\subcaption*{TR(5)}
\includegraphics[scale=1]{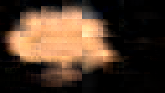}
\end{subfigure}
\begin{subfigure}[t]{0.15\textwidth}
\centering
\setlength{\abovecaptionskip}{0pt}
\setlength{\belowcaptionskip}{0pt}
\subcaption*{TMac-TT(30)}
\includegraphics[scale=1]{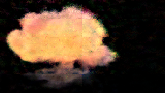}
\end{subfigure}
\begin{subfigure}[t]{0.15\textwidth}
\centering
\setlength{\abovecaptionskip}{0pt}
\setlength{\belowcaptionskip}{0pt}
\subcaption*{CP-APG(10)}
\includegraphics[scale=1]{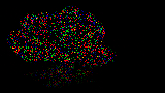}
\end{subfigure}
\begin{subfigure}[t]{0.15\textwidth}
\centering
\setlength{\abovecaptionskip}{0pt}
\setlength{\belowcaptionskip}{0pt}
\subcaption*{A-PMac-TG}
\includegraphics[scale=1]{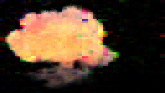}
\end{subfigure}

\begin{subfigure}[t]{0.15\textwidth}
\centering
\setlength{\abovecaptionskip}{0pt}
\setlength{\belowcaptionskip}{0pt}
\subcaption*{TG(3)}
\includegraphics[scale=1]{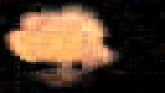}
\end{subfigure}
\begin{subfigure}[t]{0.15\textwidth}
\centering
\setlength{\abovecaptionskip}{0pt}
\setlength{\belowcaptionskip}{0pt}
\subcaption*{PMac-TG(30)}
\includegraphics[scale=1]{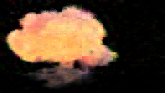}
\end{subfigure}
\begin{subfigure}[t]{0.15\textwidth}
\centering
\setlength{\abovecaptionskip}{0pt}
\setlength{\belowcaptionskip}{0pt}
\subcaption*{TR(5)}
\includegraphics[scale=1]{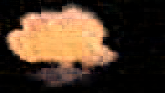}
\end{subfigure}
\begin{subfigure}[t]{0.15\textwidth}
\centering
\setlength{\abovecaptionskip}{0pt}
\setlength{\belowcaptionskip}{0pt}
\subcaption*{TMac-TT(30)}
\includegraphics[scale=1]{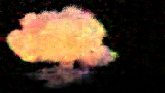}
\end{subfigure}
\begin{subfigure}[t]{0.15\textwidth}
\centering
\setlength{\abovecaptionskip}{0pt}
\setlength{\belowcaptionskip}{0pt}
\subcaption*{CP-APG(10)}
\includegraphics[scale=1]{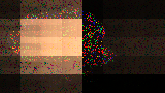}
\end{subfigure}
\begin{subfigure}[t]{0.15\textwidth}
\centering
\setlength{\abovecaptionskip}{0pt}
\setlength{\belowcaptionskip}{0pt}
\subcaption*{HaLRTC}
\includegraphics[scale=1]{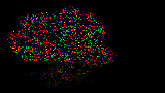}
\end{subfigure}

\begin{subfigure}[t]{0.15\textwidth}
\centering
\setlength{\abovecaptionskip}{0pt}
\setlength{\belowcaptionskip}{0pt}
\subcaption*{TG(3)}
\includegraphics[scale=1]{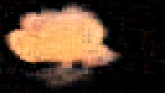}
\end{subfigure}
\begin{subfigure}[t]{0.15\textwidth}
\centering
\setlength{\abovecaptionskip}{0pt}
\setlength{\belowcaptionskip}{0pt}
\subcaption*{PMac-TG(30)}
\includegraphics[scale=1]{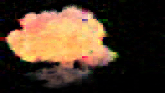}
\end{subfigure}
\begin{subfigure}[t]{0.15\textwidth}
\centering
\setlength{\abovecaptionskip}{0pt}
\setlength{\belowcaptionskip}{0pt}
\subcaption*{TR(5)}
\includegraphics[scale=1]{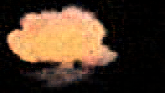}
\end{subfigure}
\begin{subfigure}[t]{0.15\textwidth}
\centering
\setlength{\abovecaptionskip}{0pt}
\setlength{\belowcaptionskip}{0pt}
\subcaption*{TMac-TT(30)}
\includegraphics[scale=1]{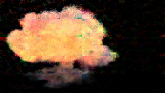}
\end{subfigure}
\begin{subfigure}[t]{0.15\textwidth}
\centering
\setlength{\abovecaptionskip}{0pt}
\setlength{\belowcaptionskip}{0pt}
\subcaption*{CP-APG(10)}
\includegraphics[scale=1]{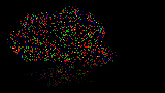}
\end{subfigure}
\begin{subfigure}[t]{0.15\textwidth}
\centering
\setlength{\abovecaptionskip}{0pt}
\setlength{\belowcaptionskip}{0pt}
\subcaption*{LRMC}
\includegraphics[scale=1]{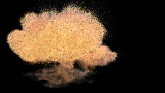}
\end{subfigure}
\caption{The completion results of the $50$-th frame of \emph{explosion} based on various algorithms, including ALS-TG, PMac-TG, A-PMac-TG, TR-ALS, TMac-TT, HaLRTC, CP-APG and LRMC. The TG ranks are $3$, $4$ and $5$. The TG factor rank are $30$, $40$ and $50$. The TR ranks are $5$, $10$ and $15$. The TT ranks are $30$, $40$ and $50$. The CP ranks are $10$, $20$ and $30$.}
\label{result_video1_figure}
\end{figure*}

Fig. \ref{result_video1_figure} shows the completion results of the $50$-th frame of \emph{explosion} and Table \ref{result_video1_table} shows the evaluation of completion for the whole video. Among the recoveries, the PMac-TG achieves the smallest relative error $0.139$ with the shortest computational time $34.32$ seconds. The A-PMac-TG does not perform as well as the PMac-TG, which indicates it may suffer from an under-fitting problem. The ALS-TG and TR-ALS algorithm are too time-consuming. The result shows our method is very efficient at large-scale tensor completion.

\begin{table}
\centering
\caption{Recovery performance of \emph{explosion}, where $10\%$ entries are observed.}
\label{result_video1_table}
\begin{tabular}{ccccc}
\toprule
Method & Rank & RE & PSNR (dB) & CPU time (s) \\
\hline
\multirow{3}{*}{ALS-TG} & 3 & 0.288 & 20.97 & $6.61\times 10^2$ \\
 & 4 & 0.253 & 22.16 & $1.87\times 10^3$ \\
 & 5 & 0.210 & 23.81 & $5.90\times 10^3$ \\
\hline
\multirow{3}{*}{PMac-TG} & 30 & 0.153 & 26.57 & $3.11\times 10^1$ \\
 & 40 & \textbf{0.139} & \textbf{27.42} & $3.43\times 10^1$ \\
 & 50 & 0.146 & 27.06 & $3.96\times 10^1$ \\
\hline
A-PMac-TG & - & 0.169 & 25.75 & $3.72\times 10^1$ \\
\hline
\multirow{3}{*}{TR-ALS} & 5 & 0.285 & 21.06 & $8.50\times 10^2$ \\
 & 10 & 0.187 & 24.82 & $5.15\times 10^3$ \\
 & 15 & 0.153 & 26.70 & $1.62\times 10^4$ \\
\hline
\multirow{3}{*}{TMac-TT} & 30 & 0.150 & 26.93 & $1.38\times 10^2$ \\
 & 40 & 0.145 & 27.26 & $2.01\times 10^2$ \\
 & 50 & 0.153 & 26.84 & $2.76\times 10^2$ \\
\hline
HaLRTC & - & 0.924 & 10.52 & $1.25\times 10^2$ \\
\hline
\multirow{3}{*}{CP-APG} & 10 & 0.949 & 10.29 & $\textbf{3.04}\times \textbf{10}^{\textbf{0}}$ \\
& 20 & 0.659 & 13.46 & $1.35\times 10^1$ \\
& 30 & 0.949 & 10.29 & $7.16\times 10^0$ \\
\hline
LRMC & - & 0.185 & 24.56 & $5.53\times 10^1$ \\
\bottomrule
\end{tabular}
\end{table}

\section{Conclusion}
\label{section_conclusion}
    
In this paper, we propose two algorithms for tensor completion using PEPS/TG representation due to its outstanding representation capability among the tensor networks. To the best of our knowledge, this is the first paper that exploits this multi-linear structure for tensor completion. The 2SDMRG algorithm is developed for initialization of TG decomposition. Based on it, alternating least squares is used to solve to the TG factors, and obtains the missing entries. In addition, with the help of parallel matrix factorization technique, we can get a fast algorithm for low rank TG approximation.  The experiments on both synthetic data and real-world color images, hyperspectral images, and monitoring videos demonstrate the superior performance of the TG decomposition based low-rank tensor completion methods.

\appendices

%



\ifCLASSOPTIONcaptionsoff
  \newpage
\fi



%



\bibliographystyle{ieeetr}
\bibliography{references_TG-ALS}

\begin{thebibliography}{10}

\bibitem{kolda2009tensor}
T.~G. Kolda and B.~W. Bader, ``Tensor decompositions and applications,'' {\em
  SIAM Review}, vol.~51, no.~3, pp.~455--500, 2009.

\bibitem{long2019low}
Z.~Long, Y.~Liu, L.~Chen, and C.~Zhu, ``Low rank tensor completion for multiway
  visual data,'' {\em Signal Processing}, vol.~155, pp.~301--316, 2019.

\bibitem{sidiropoulos2017tensor}
N.~D. Sidiropoulos, L.~De~Lathauwer, X.~Fu, K.~Huang, E.~E. Papalexakis, and
  C.~Faloutsos, ``Tensor decomposition for signal processing and machine
  learning,'' {\em IEEE Transactions on Signal Processing}, vol.~65, no.~13,
  pp.~3551--3582, 2017.

\bibitem{cichocki2015tensor}
A.~Cichocki, D.~Mandic, L.~De~Lathauwer, G.~Zhou, Q.~Zhao, C.~Caiafa, and H.~A.
  Phan, ``Tensor decompositions for signal processing applications: from
  two-way to multiway component analysis,'' {\em IEEE Signal Processing
  Magazine}, vol.~32, no.~2, pp.~145--163, 2015.

\bibitem{signoretto2011tensor}
M.~Signoretto, R.~Van~de Plas, B.~De~Moor, and J.~A. Suykens, ``Tensor versus
  matrix completion: a comparison with application to spectral data,'' {\em
  IEEE Signal Processing Letters}, vol.~18, no.~7, pp.~403--406, 2011.

\bibitem{gandy2011tensor}
S.~Gandy, B.~Recht, and I.~Yamada, ``Tensor completion and low-n-rank tensor
  recovery via convex optimization,'' {\em Inverse Problems}, vol.~27, no.~2,
  p.~025010, 2011.

\bibitem{duarte2011kronecker}
M.~F. Duarte and R.~G. Baraniuk, ``Kronecker compressive sensing,'' {\em IEEE
  Transactions on Image Processing}, vol.~21, no.~2, pp.~494--504, 2011.

\bibitem{liu2012tensor}
J.~Liu, P.~Musialski, P.~Wonka, and J.~Ye, ``Tensor completion for estimating
  missing values in visual data,'' {\em IEEE transactions on pattern analysis
  and machine intelligence}, vol.~35, no.~1, pp.~208--220, 2012.

\bibitem{liu2013tensor}
J.~Liu, P.~Musialski, P.~Wonka, and J.~Ye, ``Tensor completion for estimating
  missing values in visual data,'' {\em IEEE Transactions on Pattern Analysis
  and Machine Intelligence}, vol.~35, no.~1, pp.~208--220, 2013.

\bibitem{kilmer2013third}
M.~E. Kilmer, K.~Braman, N.~Hao, and R.~C. Hoover, ``Third-order tensors as
  operators on matrices: a theoretical and computational framework with
  applications in imaging,'' {\em SIAM Journal on Matrix Analysis and
  Applications}, vol.~34, no.~1, pp.~148--172, 2013.

\bibitem{mu2014square}
C.~Mu, B.~Huang, J.~Wright, and D.~Goldfarb, ``Square deal: Lower bounds and
  improved relaxations for tensor recovery,'' in {\em International conference
  on machine learning}, pp.~73--81, 2014.

\bibitem{du2016pltd}
B.~Du, M.~Zhang, L.~Zhang, R.~Hu, and D.~Tao, ``Pltd: Patch-based low-rank
  tensor decomposition for hyperspectral images,'' {\em IEEE Transactions on
  Multimedia}, vol.~19, no.~1, pp.~67--79, 2016.

\bibitem{kasai2016online}
H.~Kasai, ``Online low-rank tensor subspace tracking from incomplete data by cp
  decomposition using recursive least squares,'' in {\em 2016 IEEE
  International Conference on Acoustics, Speech and Signal Processing
  (ICASSP)}, pp.~2519--2523, IEEE, 2016.

\bibitem{bengua2017efficient}
J.~A. Bengua, H.~N. Phien, H.~D. Tuan, and M.~N. Do, ``Efficient tensor
  completion for color image and video recovery: low-rank tensor train,'' {\em
  IEEE Transactions on Image Processing}, vol.~26, no.~5, pp.~2466--2479, 2017.

\bibitem{xiong2018field}
B.~Xiong, Q.~Liu, J.~Xiong, S.~Li, S.~Wang, and D.~Liang, ``Field-of-experts
  filters guided tensor completion,'' {\em IEEE Transactions on Multimedia},
  vol.~20, no.~9, pp.~2316--2329, 2018.

\bibitem{he2019total}
W.~He, L.~Yuan, and N.~Yokoya, ``Total-variation-regularized tensor ring
  completion for remote sensing image reconstruction,'' in {\em ICASSP
  2019-2019 IEEE International Conference on Acoustics, Speech and Signal
  Processing (ICASSP)}, pp.~8603--8607, IEEE, 2019.

\bibitem{yang2015rank}
Y.~Yang, Y.~Feng, and J.~A. Suykens, ``A rank-one tensor updating algorithm for
  tensor completion,'' {\em IEEE Signal Processing Letters}, vol.~22, no.~10,
  pp.~1633--1637, 2015.

\bibitem{liu2015trace}
Y.~Liu, F.~Shang, L.~Jiao, J.~Cheng, and H.~Cheng, ``Trace norm regularized
  {CANDECOMP}/{PARAFAC} decomposition with missing data,'' {\em IEEE
  Transactions on Cybernetics}, vol.~45, no.~11, pp.~2437--2448, 2015.

\bibitem{zhao2015bayesian}
Q.~Zhao, L.~Zhang, and A.~Cichocki, ``Bayesian {CP} factorization of incomplete
  tensors with automatic rank determination,'' {\em IEEE Transactions on
  Pattern Analysis and Machine Intelligence}, vol.~37, no.~9, pp.~1751--1763,
  2015.

\bibitem{zhou2019tensor}
M.~Zhou, Y.~Liu, Z.~Long, L.~Chen, and C.~Zhu, ``Tensor rank learning in {CP}
  decomposition via convolutional neural network,'' {\em Signal Processing:
  Image Communication}, vol.~73, pp.~12--21, 2019.

\bibitem{liu2019low}
Y.~Liu, Z.~Long, H.~Huang, and C.~Zhu, ``Low cp rank and tucker rank tensor
  completion for estimating missing components in image data,'' {\em IEEE
  Transactions on Circuits and Systems for Video Technology}, 2019.

\bibitem{cichocki2016tensor}
A.~Cichocki, N.~Lee, I.~Oseledets, A.-H. Phan, Q.~Zhao, D.~P. Mandic, {\em
  et~al.}, ``Tensor networks for dimensionality reduction and large-scale
  optimization: part 1 low-rank tensor decompositions,'' {\em Foundations and
  Trends{\textregistered} in Machine Learning}, vol.~9, no.~4-5, pp.~249--429,
  2016.

\bibitem{xu2013parallel}
Y.~Xu, R.~Hao, W.~Yin, and Z.~Su, ``Parallel matrix factorization for low-rank
  tensor completion,'' {\em Inverse Problems and Imaging}, vol.~9, no.~2,
  pp.~601--624, 2015.

\bibitem{liu2016generalized}
Y.~Liu, F.~Shang, W.~Fan, J.~Cheng, and H.~Cheng, ``Generalized higher order
  orthogonal iteration for tensor learning and decomposition,'' {\em IEEE
  Transactions on Neural Networks and Learning Systems}, vol.~27, no.~12,
  pp.~2551--2563, 2016.

\bibitem{yang2016iterative}
L.~Yang, J.~Fang, H.~Li, and B.~Zeng, ``An iterative reweighted method for
  {Tucker} decomposition of incomplete tensors,'' {\em IEEE Transactions on
  Signal Processing}, vol.~64, no.~18, pp.~4817--4829, 2016.

\bibitem{zhang2014novel}
Z.~Zhang, G.~Ely, S.~Aeron, N.~Hao, and M.~Kilmer, ``Novel methods for
  multilinear data completion and de-noising based on tensor-{SVD},'' in {\em
  Proceedings of the IEEE Conference on Computer Vision and Pattern
  Recognition}, pp.~3842--3849, 2014.

\bibitem{zhang2017exact}
Z.~Zhang and S.~Aeron, ``Exact tensor completion using t-{SVD},'' {\em IEEE
  Transactions on Signal Processing}, vol.~65, no.~6, pp.~1511--1526, 2017.

\bibitem{oseledets2011tensor}
I.~V. Oseledets, ``Tensor-train decomposition,'' {\em SIAM Journal on
  Scientific Computing}, vol.~33, no.~5, pp.~2295--2317, 2011.

\bibitem{wang2016tensor}
W.~Wang, V.~Aggarwal, and S.~Aeron, ``Tensor completion by alternating
  minimization under the tensor train ({TT}) model,'' {\em arXiv preprint
  arXiv:1609.05587}, 2016.

\bibitem{zhao2019learning}
Q.~Zhao, M.~Sugiyama, L.~Yuan, and A.~Cichocki, ``Learning efficient tensor
  representations with ring-structured networks,'' in {\em ICASSP 2019-2019
  IEEE International Conference on Acoustics, Speech and Signal Processing
  (ICASSP)}, pp.~8608--8612, IEEE, 2019.

\bibitem{wang2017efficient}
W.~Wang, V.~Aggarwal, and S.~Aeron, ``Efficient low rank tensor ring
  completion,'' in {\em Computer Vision (ICCV), 2017 IEEE International
  Conference on}, IEEE, 2017.

\bibitem{huang2019provable}
H.~Huang, Y.~Liu, and C.~Zhu, ``Provable model for tensor ring completion,''
  {\em arXiv preprint arXiv:1903.03315}, 2019.

\bibitem{ye2018tensor}
K.~Ye and L.-H. Lim, ``Tensor network ranks,'' {\em arXiv preprint
  arXiv:1801.02662}, 2018.

\bibitem{bridgeman2017hand}
J.~C. Bridgeman and C.~T. Chubb, ``Hand-waving and interpretive dance: an
  introductory course on tensor networks,'' {\em Journal of Physics A:
  Mathematical and Theoretical}, vol.~50, no.~22, p.~223001, 2017.

\bibitem{pivzorn2011time}
I.~Pi{\v{z}}orn, L.~Wang, and F.~Verstraete, ``Time evolution of projected
  entangled pair states in the single-layer picture,'' {\em Physical Review A},
  vol.~83, no.~5, p.~052321, 2011.

\bibitem{lubasch2014algorithms}
M.~Lubasch, J.~I. Cirac, and M.-C. Banuls, ``Algorithms for finite projected
  entangled pair states,'' {\em Physical Review B}, vol.~90, no.~6, p.~064425,
  2014.

\bibitem{verstraete2008matrix}
F.~Verstraete, V.~Murg, and J.~I. Cirac, ``Matrix product states, projected
  entangled pair states, and variational renormalization group methods for
  quantum spin systems,'' {\em Advances in Physics}, vol.~57, no.~2,
  pp.~143--224, 2008.

\bibitem{tan2014tensor}
H.~Tan, B.~Cheng, W.~Wang, Y.-J. Zhang, and B.~Ran, ``Tensor completion via a
  multi-linear low-n-rank factorization model,'' {\em Neurocomputing},
  vol.~133, pp.~161--169, 2014.

\bibitem{xu2013block}
Y.~Xu and W.~Yin, ``A block coordinate descent method for regularized
  multiconvex optimization with applications to nonnegative tensor
  factorization and completion,'' {\em SIAM Journal on Imaging Sciences},
  vol.~6, no.~3, pp.~1758--1789, 2013.

\bibitem{zhu2014spectral}
F.~Zhu, Y.~Wang, B.~Fan, S.~Xiang, G.~Meng, and C.~Pan, ``Spectral unmixing via
  data-guided sparsity,'' {\em IEEE Transactions on Image Processing}, vol.~23,
  no.~12, pp.~5412--5427, 2014.

\bibitem{zhu2014structured}
F.~Zhu, Y.~Wang, S.~Xiang, B.~Fan, and C.~Pan, ``Structured sparse method for
  hyperspectral unmixing,'' {\em ISPRS Journal of Photogrammetry and Remote
  Sensing}, vol.~88, pp.~101--118, 2014.

\end{thebibliography}

%








\end{document}